\documentclass[twoside]{article}

\usepackage[accepted]{arxiv}
\usepackage[round]{natbib}

\usepackage{multicol}
\usepackage{multirow}

\usepackage{caption}
\usepackage{graphicx}

\usepackage{algorithm}
\usepackage{algorithmic}
\usepackage{subfigure}

\usepackage{amsmath}
\usepackage{amssymb}
\usepackage{hyperref}

\usepackage{amsthm}
\usepackage{mathtools}

\usepackage[capitalise]{cleveref}
\usepackage{booktabs}

\usepackage{comment}
\usepackage{setspace}

\usepackage{stmaryrd} 

\newcommand{\p}[0]{\mathbb{P}}
\newcommand{\I}[0]{\mathbb{I}}
\newcommand{\E}[0]{\mathbb{E}}
\newcommand{\R}[0]{\mathbb{R}}
\newcommand{\N}[0]{\mathbb{N}}
\newcommand{\X}[0]{\mathcal{X}}

\usepackage{xspace}
\def\roc{{\rm ROC\xspace}}
\def\auc{{\rm AUC\xspace}}

\usepackage{lipsum}  
\usepackage{mathabx}

\newtheorem{theorem}{Theorem}
\newtheorem{definition}{Definition}
\newtheorem{lemma}{Lemma}
\newtheorem{proposition}{Proposition}

\newtheorem{remark}{Remark}
\newtheorem{example}{Example}
\newtheorem{assumption}{Assumption}

\DeclarePairedDelimiter\abs{\lvert}{\rvert}%
\DeclarePairedDelimiter\norm{\lVert}{\rVert}%

\makeatletter
\let\oldabs\abs
\def\abs{\@ifstar{\oldabs}{\oldabs*}}
\let\oldnorm\norm
\def\norm{\@ifstar{\oldnorm}{\oldnorm*}}
\makeatother

\DeclareMathOperator*{\sign}{sign}

\usepackage{color}
\definecolor{aoenglish}{rgb}{0.0,0.5,0.0}
\definecolor{gray}{rgb}{0.7,0.7,0.7}

\begin{document}

%
\runningtitle{Learning Fair Scoring Functions}

%

\twocolumn[

\aistatstitle{Learning Fair Scoring Functions:\\ Bipartite Ranking under
ROC-based Fairness Constraints}

\aistatsauthor{ Robin Vogel \And Aurélien Bellet \And  Stephan Clémençon}

\aistatsaddress{LTCI, T\'el\'ecom Paris,\\Institut Polytechnique de Paris, France
    \And INRIA, France
    \And LTCI, T\'el\'ecom Paris,\\Institut Polytechnique de Paris, France}
]


\begin{abstract}

Many applications of AI involve \textit{scoring} individuals
using a learned function of their attributes. These predictive risk scores are
then used to take decisions based on whether the score exceeds a certain
threshold, which may vary depending on the context.
The level of delegation granted to such systems in critical applications
like credit lending and medical diagnosis will heavily depend on how
questions of \textit{fairness} can be answered.
In this paper, we study fairness for the problem of learning scoring functions
from binary labeled data, a classic learning task known as  \textit{bipartite
ranking}. We argue that the functional nature of the $\roc$ curve, the gold
standard measure of ranking accuracy in this context, leads to several ways
of formulating fairness constraints. We introduce general families of fairness
definitions based on the $\auc$ and on $\roc$ curves, and show that our
$\roc$-based constraints can be instantiated such that classifiers
obtained by thresholding
the scoring function
satisfy classification fairness for a desired range of
thresholds.
We establish generalization
bounds for scoring functions learned under such constraints, design practical
learning algorithms and show the relevance our approach
with numerical
experiments on real and synthetic data.

\end{abstract}

\section{INTRODUCTION}\label{sec:introduction}

With the availability of data at ever finer granularity
and the development of technological bricks to efficiently store and
process this data, the infatuation with machine learning (ML) and artificial
intelligence (AI) is spreading to nearly all fields (science, transportation,
energy, medicine, security, banking, insurance, commerce...).
Expectations are high.
There is no denying
the opportunities, and we can rightfully hope for an increasing number of
successful deployments in the near future.
However, AI will keep its promises only if certain issues are addressed. In
particular, ML systems that make significant decisions for
humans, regarding for instance credit lending in the banking sector 
\citep{Chen2018}, diagnosis
in medicine \citep{Deo15} or recidivism prediction in
criminal justice \citep{rudin2018age}, should guarantee that they do not
penalize certain groups of individuals.

Hence, stimulated by the societal expectations, notions of \textit{fairness}
in ML as well guarantees that they can be fulfilled by models trained under
appropriate constraints have recently been the
subject of a good deal of attention in the literature, see \textit{e.g.} 
\citep{Dwork12,Kleinberg17} among others. Fairness constraints are
generally modeled by means of a (qualitative) \textit{sensitive variable},
indicating membership to a certain group (\textit{e.g.}, ethnicity, gender).
The vast majority of the work dedicated to algorithmic fairness in ML focuses on binary classification.
In this context, fairness constraints force classifiers to have similar
true positive rates (or false positive rates)
across sensitive groups. For instance, \cite{Hardt16,Pleiss17} propose to modify a pre-trained classifier in order to
fulfill such constraints without deteriorating too much the classification
performance.
Other work incorporates fairness constraints in the learning stage 
\citep[see \emph{e.g.},][]{Agarwal17,Woodworth17,Zafar17,Zafar17b,Zafar2019,Menon18,Bechavod18}.
In addition to algorithms, statistical guarantees (in the form of
generalization bounds) are crucial for fair ML, as they ensure
that the desired fairness constraint will be met at deployment. Such learning
guarantees
have been established by \cite{Donini2018} for the case of fair
classification.

Many real-world problems are however not concerned with learning a binary
classifier but rather aim to learn a \emph{scoring function}.
This statistical learning problem is known as 
\textit{bipartite ranking} and covers in particular tasks such as credit
scoring in banking, pathology scoring in medicine or recidivism scoring in criminal justice, for
which fairness is a major concern \citep{NIPS2019_8604}.
While it can be formulated in the same probabilistic framework as binary
classification, bipartite ranking is not a local learning problem: 
the goal is not to guess whether a binary label $Y$ is positive or
negative from an input observation $X$ but to rank any collection of
observations $X_1,\; \ldots,\; X_n$ by means of a scoring function $s:
\mathcal{X}\rightarrow \mathbb{R}$ so that observations with positive labels
are ranked higher with large probability. Due to the global nature of the
task, evaluating the performance is itself a challenge. The gold standard
measure, the $\roc$ curve, is functional: it is the PP-plot of the false
positive rate (FPR) \textit{vs} the true positive rate (TPR), and
the higher the curve, the more accurate the ranking induced by $s$.
Sup-norm optimization of the $\roc$ curve has been investigated by 
\cite{CV09ieee,CV09CA}, while most of the literature
focuses on the maximization of scalar summaries of the $\roc$ curve such as
the $\auc$ criterion \citep{AGHHPR05,Clemencon08Ranking,Zhao2011a} or
alternative measures \citep{Rud06,Clemencon2007,menon2016bipartite}.

A key advantage of learning a scoring function over learning a classifier is
the flexibility in thresholding the scores so as to obtain false/true positive
rates that fit the particular operational constraints in which the decision is
taken.
A natural fairness requirement in this context is that a fair scoring
function should lead to fair decisions \emph{for all thresholds of interest}.
To help fix ideas and grasp the methodological challenge, we describe below a
concrete example to motivate our work.

\begin{example}[Credit-risk screening]\label{ex:credit}
A bank grants a loan to a client with
socio-economic features $X$ if his/her score $s(X)$ is above a certain
threshold $t$.
As the degree of risk aversion of the bank may vary, the precise
deployment threshold $t$ is unknown when choosing the scoring function $s$,
although the bank is generally interested in regimes where the
probability of default is sufficiently small (low FPR). The bank
would like to design a scoring function that ranks higher the clients
that are more likely to repay the loan (\emph{ranking performance}), while
ensuring that any threshold in the regime of interest will lead to similar
false negative rates across sensitive groups (\emph{fairness constraint}).
\end{example}

\begin{figure}[t]
    \centering
    \includegraphics[width=0.95\columnwidth]{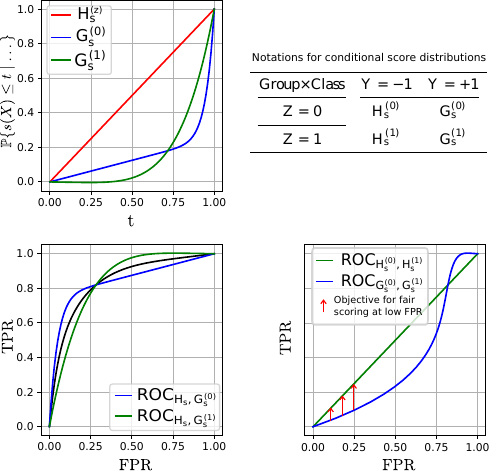}
    \caption{Illustrating the limitations of $\auc$-based fairness.
    Here, the group-wise positive/negative distributions (top)
    satisfy $\auc_{H_s, G_s^{(0)}} = \auc_{H_s, G_s^{(1)}}$ (bottom left),
    but yield very different TPR's at low FPR's (bottom left).
    Our new $\roc$-based constraints can align scores distributions
    where it matters, \textit{e.g.} for low FPR's as in \cref{ex:credit} 
    (bottom right).
    }
    \label{fig:example-1}
\end{figure}

\textbf{Contributions.}
In this work, we provide a thorough study of fairness in bipartite ranking.
Our starting point is a number of fairness measures
introduced independently in recent papers from different communities 
\citep{Borkan2019,Beutel2019FRR,NIPS2019_8604}. We first show that these are
special cases of a general family of fairness constraints based on the $\auc$,
that we precisely characterize. We then argue that, because it is defined from
scalar summaries of functional curves, $\auc$-based fairness is oblivious to
potentially large disparities between groups at particular locations of
the score distribution (see \cref{fig:example-1}, bottom left). As a
consequence,
they fail to
address use-cases where fairness is needed at specific thresholds
(as in Example~\ref{ex:credit}).
To overcome these limitations, we introduce a
novel functional view
of fairness based on $\roc$ curves. These richer
\emph{pointwise $\roc$-based constraints} can be instantiated to align
group-wise score distributions at specific functional points (see 
\cref{fig:example-1}, bottom right) and thereby ensure that
classifiers obtained by thresholding the scoring function
satisfy classification fairness for a certain range of thresholds,
as desired in cases like 
Example~\ref{ex:credit}.

Based on the above, we then introduce
empirical risk minimization formulations for learning fair scoring
functions under both $\auc$ and $\roc$-based fairness constraints and
establish the first generalization bounds for fair bipartite ranking.
Due to the complex nature of the ranking measures,
our proof techniques largely differ from the classification results of
\cite{Donini2018} as they require
non standard technical tools
(\emph{e.g.} to control deviations of ratios of $U$-statistics).
In addition to our conceptual contributions and theoretical analysis, we
propose efficient training algorithms based on gradient descent and illustrate
the practical relevance of our approach on synthetic and real datasets.

\textbf{Outline.}
The paper is organized as follows. \cref{sec:preliminaries} reviews bipartite
ranking as well as existing fairness notions for classification and ranking.
\cref{sec:fairness_ranking} studies $\auc$-based fairness constraints and
propose richer $\roc$-based constraints.
In Section~\ref{sec:beyond_auc_cons}, we formulate the problem of fair scoring under both $\auc$ and $\roc$-based fairness constraints
and prove statistical learning guarantees.
\cref{sec:experiments} presents numerical experiments, and we conclude in 
\cref{sec:conclusion}.
Due to space limitations, some technical details and additional experiments
are postponed to the supplementary.


\section{BACKGROUND \& RELATED WORK}\label{sec:preliminaries}

In this section, we introduce the main concepts involved in the
subsequent analysis and review related work.
Here and throughout, the indicator function of any event $\mathcal{E}$ is
denoted by $\I\{\mathcal{E}\}$ and the pseudo-inverse of any cumulative
distribution function (c.d.f.) function $F: \R \to [0,1]$ by $F^{-1}(u) =
\inf\left\{ t \in \R : F(t) \ge u \right\}$.

\subsection{Probabilistic Framework}

Let $X$ and $Y$ be two random variables: $Y$ denotes the
binary output label (taking values in $\{-1, +1\}$) and $X$ denotes the input
features, taking values in a space $\X \subset \R^d$ with $d \ge 1$
and modeling some information hopefully useful to predict $Y$. 
For convenience, we introduce the proportion of positive instances $p := \p\{ Y
= +1 \}$, as well as $G$ and $H$, the conditional distributions of $X$ given
$Y=+1$ and $Y=-1$ respectively. The joint distribution of $(X,Y)$ is fully
determined by the triplet $(p, G, H)$.
Another way to specify the distribution of $(X,Y)$ is through the pair $
(\mu,\eta)$ where $\mu$ denotes the marginal distribution of $X$ and $\eta$
the function $\eta(x) := \p\{ Y=+1 \mid X=x \}$. With these
notations, one may write $\eta(x) = p (dG/dH)(x)/(1-p + p(dG/dH)(x))$ and $\mu = pG + (1-p)H$.

In the context of fairness, we consider a third random variable $Z$ which
denotes the sensitive attribute taking values in $\{0, 1\}$.
The pair $(X,Y)$ is said to belong to salient group $0$ (resp. $1$) when $Z=0$
(resp. $Z=1$).
The distribution of the triplet $(X, Y, Z)$ can be expressed as a mixture of
the distributions of $X, Y | Z=z$. Following the conventions described above,
we introduce the quantities $p_z, G^{(z)}, H^{(z)}$
as well as $\mu^{(z)}, \eta^{(z)}$. For
instance, $p_0 = \p\{ Y= +1 | Z = 0 \}$ and the distribution of $X | Y=+1, Z=0$
is written $G^{(0)}$, \textit{i.e.} for $A \subset \X$,
$G^{(0)}(A) = \p\{ X\in A | Y=+1, Z=0 \}$.
We denote the probability of belonging to group $z$ by $q_z:=\p \{Z=z\}$,
with $q_0 = 1 - q_1$.


\subsection{Bipartite Ranking}\label{subsec:bipartite-ranking}

The goal of bipartite ranking is to learn an order relationship on $\X$
for which positive instances are ranked higher than negative ones with high probability.
This order is defined by transporting the natural order on the real line to
the feature space through a scoring function $s: \X \to\R$.
Given a distribution $F$ over $\X$ and a scoring function $s$, we denote by
$F_s$ the cumulative distribution function of $s(X)$ when $X$ follows
$F$.
Specifically:
\begin{align*}
    G_s(t) & := \p\{ s(X) \le t \mid Y = +1 \}
    = G(s(X) \le t) ,\\
    H_s(t) & := \p\{ s(X) \le t \mid Y = -1\}
    = H(s(X) \le t).
\end{align*}
\textbf{ROC analysis.}
ROC curves are widely used to visualize 
the dissimilarity between two
real-valued distributions in many applications, \textit{e.g.}
anomaly detection, medical diagnosis, information retrieval. 
\begin{definition}[ROC curve]
\label{def:roc-curve}
    Let $g$ and $h$ be two cumulative distribution functions on $\R$.
    The $\roc$ curve related to $g$ and $h$ is the graph of the mapping:
    \begin{align*}
	\roc_{h,g} : \alpha \in [0,1] \mapsto 1 - g \circ h^{-1}
        (1-\alpha).
    \end{align*}
    When $g$ and $h$ are continuous, it can alternatively be defined
    as the parametric curve $t \in \R \mapsto (1-h(t), 1-g(t))$.
\end{definition}
The classic area under the $\roc$ curve ($\auc$) criterion is
a scalar summary of the functional measure of dissimilarity $\roc$.
Formally, we have:
\begin{align*}
    \auc_{h,g} & := \int \roc_{h,g}( \alpha) d\alpha
       = \p \{ S > S' \} + \frac{1}{2} \p\{ S=S' \}, 
\end{align*} 
where $S$ and $S'$ denote independent random variables,
whose c.d.f.'s are $h$ and $g$ respectively.

In 
bipartite ranking, one focuses on 
the ability of the
scoring function $s$ to separate positive from negative data. This is
reflected by  $\roc_{H_s, G_s}$, which gives the false positive rate 
\emph{vs.}
true positive rate 
of binary classifiers $g_{s, t} : x
\mapsto 2\cdot\I\{s
(x) > t \} - 1$ obtained by thresholding $s$ at all possible
thresholds $t \in \R$.
The global summary $\auc_{H_s, G_s}$ serves as a
standard performance measure \citep{Clemencon08Ranking}.


\textbf{Empirical estimates.}
In practice, the scoring function $s$ is learned based on a training set $\{
(X_i, Y_i) \}_{i=1}^n$ of $n$ i.i.d. copies of the random pair $(X,Y)$.
Let $n_+$ and $n_-$ be the number of positive and negative data points
respectively.
We introduce $
\widehat{G}_s$ and $\widehat{H}_s$, the empirical counterparts of $G_s$ and
$H_s$:
\begin{align*}
    \widehat{G}_s(t) &:= (1/n_+) \textstyle\sum_{i=1}^n \I\{ Y_i = +1, s(X_i)
    \le t \}, \\
    \widehat{H}_s(t) &:= (1/n_-) \textstyle\sum_{i=1}^n \I\{ Y_i = -1, s(X_i)
    \le t
    \}.
\end{align*} 
Note that the denominators $n_+$ and $n_-$ are sums of i.i.d. random (indicator)
variables.
For any two distributions $F,F'$ over $\mathbb{R}$, we denote the empirical
counterparts of $\auc_{F,F'}$ and $\roc_{F,F'}$ by $
\widehat{\auc}_
{F,F'}
        := \auc_{\widehat{F},\widehat{F'}}$ and
    $\widehat{\roc}_{F,F'}(\cdot)
        := \roc_{\widehat{F},\widehat{F'}}(\cdot)$ respectively. In
        particular, we
        have:
\begin{align*}
    \widehat{\auc}_{H_s,G_s} 
    := & ~\textstyle\frac{1}{n_+n_-} \sum_{i < j} 
    K( (s(X_i), Y_i), (s(X_j), Y_j)),
\end{align*} 
where $K( (t, y), (t', y')) = \I\{ (y-y')(t-t') > 0\} + \I\{ y \ne y', t=t' \}/2$ for any $t,t' \in \R^2, y,y' \in \{-1,+1\}^2$.
Empirical risk minimization for bipartite ranking typically
consists in maximizing $\widehat{\auc}_{H_s,G_s}$ over a class of scoring
functions \citep[see \emph{e.g.}][]{Clemencon08Ranking,Zhao2011a}.

\subsection{Fairness in Binary Classification} 

In binary classification, the goal is to
learn a mapping function $g: \X \mapsto \{-1, +1\}$ that predicts the output
label $Y$ from the input random variable $X$ as accurately as possible (as
measured by an appropriate loss function).
Any classifier $g$ can be defined by its unique acceptance set $A_g := \{ x \in
\X \mid g(x) = +1 \} \subset \X$.
%

Existing notions of fairness for binary classification
\citep[see][for a detailed treatment]{Zafar2019} aim to
ensure that $g$ makes similar predictions (or errors) for the two groups. We
mention here the common
fairness definitions that depend on the ground truth label
$Y$. \emph{Parity in mistreatment} requires that the proportion of errors is
the same for the two groups:
\begin{align}
    M^{(0)}(g) = M^{(1)}(g),
    \label{eq:parity-in-mistreatment}
\end{align}
where $M^{(z)}(g) := \p\{ g(X) \ne Y \mid Z=z \}$.
While this requirement is natural, it considers that all errors are equal: in
particular, one can have a high
false positive rate (FPR) $H^{(1)}(A_g)$ for one group and a high false
negative
rate (FNR) $G^{(0)}(A_g)$ for the other. This can be considered unfair when
acceptance is an advantage, \emph{e.g.} being granted a loan in
Example~\ref{ex:credit}). A solution
is to consider \emph{parity in false positive rates} and/or 
\emph{parity in false negative rates}, which respectively write:
\begin{align}
    H^{(0)}(A_g) = H^{(1)}(A_g)\text{ and } G^{(0)}(A_g) = G^{(1)}(A_g).
    \label{eq:parity-in-fpr-fnr}
\end{align}

\begin{remark}[Connection to bipartite ranking]\label{remark:connection-bipartite-ranking}
A score function $s: \X\to\R$ induces an infinite collection of binary classifiers $g_{s, t} : x \mapsto 2\cdot\I\{ s
(x) > t \} - 1$.
While one could fix a threshold $t\in\R$ in advance and enforce fairness on
$g_
{s,t}$,
we are interested here in notions of fairness for the
score function itself (see Example~\ref{ex:credit}).
\end{remark}

\subsection{Fairness in Ranking}\label{subsec:auc_cons_for_fairness_in_ranking}

Fairness for rankings has been mostly considered in
the informational retrieval and
recommender systems communities. Given a set of items with \emph{known
relevance scores}, they aim to extract a (partial) ranking that balances
utility and notions of fairness at the group or individual level, or through a
notion of exposure over several queries 
\citep{Zehlike2017FAIRAF,CelisSV17,Biega2018,Singh2018}.
\cite{Singh2019} and \cite{Beutel2019FRR} extend the above work to the 
\emph{learning to rank} framework, where the task is to learn relevance scores
and ranking policies from a certain number of observed \emph{queries} that
consist of query-item features and item relevance scores.
This is fundamentally different from the bipartite ranking setting considered
here.



\textbf{AUC constraints.}
In a setting closer to ours, 
\cite{NIPS2019_8604} introduce measures to quantify the
fairness of a known scoring function on binary labeled data (they do not
address learning).
Their approach is based on the $\auc$,
which can be seen as a measure of homogeneity between
distributions \citep{ClemenconNIPS2009}.
Similar definitions of fairness are also considered in 
\citep{Beutel2019FRR,Borkan2019}.

Introduce $G_s^{(z)}$ (resp. $H_s^{(z)}$) as the c.d.f. of the score on the positives
(resp.  negatives) of group $z\in\{0,1\}$, \textit{i.e.}
$G_s^{(z)} (t) = G^{(z)}(s(X) \le t )$ and
$H_s^{(z)}(t)  = H^{(z)} (s(X) \le t )$,
for any $t\in\R$.
Precise examples of $\auc$-based fairness constraints
include: 1) the \textit{intra-group pairwise $\auc$ fairness} \citep{Beutel2019FRR},
\begin{align}\label{rk-cons:intra-pairwise} 
    \auc_{H_s^{(0)}, G_s^{(0)}} = \auc_{H_s^{(1)}, G_s^{(1)}},
\end{align}
which requires the ranking performance to be equal \emph{within} groups,
2) the \textit{Background Negative Subgroup Positive (BNSP) $\auc$ fairness} 
\citep{Borkan2019},
\begin{align}\label{rk-cons:pairwise-accuracy} 
    \auc_{ H_s, G^{(0)}_s } &= \auc_{ H_s, G^{(1)}_s },
\end{align}
which enforces that positive instances from either group have the same
probability of being ranked higher than a negative example,
3) the \textit{inter-group pairwise $\auc$ fairness} \citep{NIPS2019_8604},
\begin{align}\label{rk-cons:inter-group-pairwise-fairness} 
    \auc_{H_s^{(0)}, G_s^{(1)}} = \auc_{H_s^{(1)}, G_s^{(0)}},
\end{align}
which imposes that the positives of a group can be distinguished
from the negatives of the other group as effectively for both groups.
Many more $\auc$-based fairness constraints are possible:
we give examples (some of them novel) in the supplementary material.


\section{FROM AUC TO ROC-BASED FAIRNESS CONSTRAINTS}
\label{sec:fairness_ranking}


In this section, we first provide a new general framework
to characterize all relevant $\auc$ constraints. We then highlight some
limitations of $\auc$
fairness constraints,
which serve as motivation to introduce our richer
\emph{pointwise ROC-based fairness constraints}.


\subsection{A Family of AUC Fairness Constraints}
\label{subsec:general-auc-fairness-formulation}


All proposed $\auc$-based fairness constraints in the literature
follow a common structure, which we precisely characterize.

Denote by $(e_1, e_2, e_3, e_4)$ the canonical basis of $\R^4$,
as well as by $\mathbf{1}$ the constant vector $\mathbf{1} = \sum_{k=1}^4 e_k$.
$\auc$ constraints are expressed in the form of equalities
of the $\auc$'s between mixtures of the c.d.f.'s $D(s)$, with:
    $ D(s) := (H^{(0)}_s, H^{(1)}_s,
    G^{(0)}_s, G^{(1)}_s)^\top$.
Formally, introducing the probability vectors $\alpha, \beta, \alpha', \beta' \in 
\mathcal{P}$ where
$\mathcal{P} = \{ v \mid v \in \R_+^4, \mathbf{1}^\top v = 1 \}$,
they write as:
\begin{align}\label{eq:fairness-mixture}
    \auc_{\alpha^\top D(s), \beta^\top D(s)} = \auc_{\alpha'^\top D(s),
    \beta'^\top D(s)}.
\end{align}
However, observe that \cref{eq:fairness-mixture} is under-specified in the
sense that
it includes constraints that actually give an advantage to one of the groups.

We thus introduce a general framework to formulate all \emph{relevant}
AUC-based constraints (and only those) 
as a linear combination of 5 elementary constraints.
Given a scoring function $s$, let the vector
    $C(s) = (C_1(s), \dots, C_5(s))^\top$
where the $C_l(s)$'s are elementary fairness measurements.
Specifically, the value of $|C_1(s)|$ (resp. $|C_2(s)|$) quantifies
the
resemblance of the
distribution of the negatives (resp.
positives) between the two sensitive attributes:
\begin{align*}
    C_1(s) &= \auc_{H^{(0)}_s, H^{(1)}_s} - 1/2,\\
    C_2(s) &=   1/2 - \auc_{G^{(0)}_s, G^{(1)}_s},
\end{align*}
while $C_3(s)$, $C_4(s)$ and $C_5(s)$ measure the 
difference in ability of a score to discriminate between positives and negatives
for any two pairs of sensitive attributes:
\begin{align*}
    C_3(s) &= \auc_{H^{(0)}_s, G^{(0)}_s} - \auc_{H^{(0)}_s, G^{(1)}_s}, \\
    C_4(s) &= \auc_{H^{(0)}_s, G^{(1)}_s} - \auc_{H^{(1)}_s, G^{(0)}_s}, \\
    C_5(s) &= \auc_{H^{(1)}_s, G^{(0)}_s} - \auc_{H^{(1)}_s, G^{(1)}_s}.
\end{align*}
The family of fairness constraints we consider is then the set of linear
combinations of the $C_l(s) = 0$:
\begin{align}\label{eq:barycenter-constraint-formulation}
    \mathcal{C}_\Gamma(s): \quad \Gamma^\top C(s) &= \textstyle\sum_{l=1}^5
    \Gamma_l C_l (s) = 0,
\end{align}
where $\Gamma = (\Gamma_1 , \dots, \Gamma_5)^\top \in \R^5$.

\begin{theorem}\label{th:generality_fairness_constraint}
    The following statements are equivalent:
    \begin{enumerate}\setlength\itemsep{.0em}
    \item 
        \cref{eq:fairness-mixture} is satisfied for any measurable scoring
        function $s$
        when $H^{(0)} = H^{(1)}$, $G^{(0)} = G^{(1)}$ 
        and $\mu(\eta(X) = p) < 1$,
    \item \cref{eq:fairness-mixture} is equivalent to
        $\mathcal{C}_\Gamma(s)$ for some $\Gamma \in \R^5$,
    \item $(e_1 + e_2)^\top [ (\alpha - \alpha') - (\beta-\beta')] = 0$.
    \end{enumerate}
\end{theorem}

Theorem~\ref{th:generality_fairness_constraint} shows that our general family
defined by \cref{eq:barycenter-constraint-formulation} compactly captures all
relevant AUC-based fairness constraints
\citep[including those proposed by][]{Beutel2019FRR,Borkan2019,NIPS2019_8604}
while ruling out the ones that are not
satisfied when $H^{(0)} = H^{(1)}$ and $G^{(0)} = G^ {(1)}$ (which are in fact
\emph{unfairness} constraints). 
Their parameters $\Gamma$ are provided in \cref{tab:table_lambdas_main}.
We refer to the supplementary for the
proof of this result and examples of novel fairness constraints that can be
expressed with \cref{eq:barycenter-constraint-formulation}.

\begin{table*}[t]
    \centering
    \caption{Value of $\Gamma$ in our formulation of
    \cref{eq:barycenter-constraint-formulation} for AUC-based constraints
    introduced in previous work.}\label{tab:table_lambdas_main}
    {
    \renewcommand{\arraystretch}{1.25}
    {\footnotesize
    \begin{tabular}[h]{lccccc}
    \hline
    $\auc$-based fairness constraint
    & $\Gamma_1$ & $\Gamma_2$ & $\Gamma_3$ & $\Gamma_4$ & $\Gamma_5$ \\
    \hline
    Intra-group pairwise \citep{Beutel2019FRR}, subgroup $\auc$ 
    \citep{Borkan2019} & $0$ & $0$ & $\frac{1}
    {3}$ & 
    $\frac{1}{3}$ &  $\frac{1}{3}$ \\
    BNSP $\auc$ \citep{Borkan2019}, pairwise accuracy
    \citep{Beutel2019FRR}
     & $0$ & $0$ & $\frac{q_0(1-p_0)}{1-p}$ & 
    $0$ & $\frac{q_1(1-p_1)}{1-p}$ \\
    BPSN $\auc$ \citep{Borkan2019,Beutel2019FRR,NIPS2019_8604}
    & $0$ & $0$ & $\frac{q_0p_0}{2p}$ 
    & $\frac{1}{2}$ & $\frac{q_1p_1}{2p}$ \\
    Zero Average Equality Gap \citep{Borkan2019}
     & $0$ & $1$ & $0$ & $0$ & $0$ \\
    Inter-group pairwise \citep{Beutel2019FRR}, x$\auc$
    \citep{NIPS2019_8604}
    & $0$ & $0$ & $0$ 
    & $1$ & $0$ \\
    \hline
    \end{tabular}
    }
    }
\end{table*}

As we show in \cref{subsec:learning-under-auc-cons}, our unifying framework
enables the design of general formulations and statistical
guarantees for learning fair scoring functions, which can then
be instantiated to the specific notion of
AUC-based fairness that the practitioner is interested in.

\subsection{Limitations of AUC-based Constraints}\label{sec:limits-auc-based-constraints}

To illustrate the fundamental limitations of $\auc$-based fairness
constraints, we will rely on the credit-risk screening use case described in
\cref{ex:credit}. Imagine that the scoring function $s$ gives the
c.d.f.'s
$H_s^{(z)}$ and $G_s^{(z)}$ shown in \cref{fig:example-1} (top). Looking at
$G_s^{(1)}$, we can see that creditworthy ($Y = +1$) individuals of the
sensitive group $Z=1$ do not have scores smaller than $0.5$
and have an almost constant positive density of scores between 0.6 and 1.
On the other hand, the scores of creditworthy individuals of group $Z=0$
are sometimes low but are mostly concentrated around
$1$ (greater than 0.80), as seen from $G_s^{(0)}$. The distribution of scores
for individuals who do not repay their loan ($Y=-1$) is the same across
groups.

Even though the c.d.f.'s $G_s^{(0)}$ and $G_s^{(1)}$
are very different, the scoring function $s$ satisfies the $\auc$
constraint in \cref{rk-cons:pairwise-accuracy}, as can be seen from 
\cref{fig:example-1} (bottom left). This means that creditworthy individuals
from either group have the same probability of being ranked higher than a
``bad borrower''. However, using high thresholds (which correspond to low
probabilities of default on the granted loans) will lead to
unfair decisions
for one group. For instance, using $t = 0.85$ gives a FNR of
30\% for group 0 and of 60\% for group $1$, as can be seen from 
\cref{fig:example-1} (top).
If the proportion of creditworthy people is the same in each group ($p_0 q_0 =
p_1 q_1$), we would reject twice as much creditworthy people of group $1$ than
of group $0$! This is blatantly unfair in the sense of parity in FNR defined
in \cref{eq:parity-in-fpr-fnr}.

In general, fairness constraints defined by the equality between two $\auc$'s
only quantify a stochastic order between distributions,
not the equality between these distributions.
In fact, for continuous $\roc$s, the equality
between their two $\auc$'s only implies that the two $\roc$'s intersect
at some unknown point.
As a consequence, $\auc$-based fairness can only guarantee that
there exists \emph{some} threshold $t\in \R$ that
induces a non-trivial classifier $g_{s, t} : x \mapsto 2\cdot\I\{ s
(x) > t \} - 1$
satisfying a notion of fairness for classification
(see the supplementary for details).
Unfortunately, the value of $t$
and the corresponding FPR of the ROC curves are not known
in advance and are difficult to control.
For the distributions of \cref{fig:example-1},
we see that the classifier $g_{s,t}$ is fair in FNR only for $t=0.72$ (20\%
FNR for each group) but has a rather high FPR (i.e., probability of
default) of $\sim$25\%, which may be not sustainable for the bank.

\subsection{Learning with Pointwise ROC-based Fairness Constraints}
\label{sec:new-pointwise-roc}
To impose richer and more targeted fairness conditions,
we propose to use \emph{pointwise $\roc$-based fairness constraints} as an
alternative to $\auc$-based constraints. We
start from the ``ideal fairness goal'' of enforcing
the equality of the score distributions of the
positives (resp. negatives) between the two groups, \textit{i.e.}
$G_s^{(0)} = G_s^{(1)}$ (resp. $H_s^{(0)} = H_s^{(1)}$).
This strong functional criterion can be expressed in terms of ROC curves.
For $\alpha \in [0,1]$, consider the deviations between the 
\textit{positive} (resp. \textit{negative}) \textit{inter-group ROCs}
and the identity function:
\begin{align*}
    \Delta_{G, \alpha}(s) &:= \roc_{G^{(0)}_s, G^{(1)}_s}(\alpha) - \alpha, \\
    \big( \text{resp. } \Delta_{H, \alpha}(s) & := \roc_{H^{(0)}_s,H^{(1)}_s}
    (\alpha) - \alpha
    \big).
\end{align*}
The aforementioned condition of equality between the distribution of the positives
(resp. negatives) of the two groups are equivalent to satisfying
$\Delta_{G, \alpha}(s) = 0$ (resp. $\Delta_{H, \alpha}(s) = 0$)
for any $\alpha \in [0,1]$.
When both of those conditions 
are satisfied,
all of the AUC-based fairness constraints covered by 
\cref{eq:barycenter-constraint-formulation}
are verified, as it is easy to see that $C_l(s) = 0$ for all $l \in \{1,
\dots,
5\}$. Furthermore,
guarantees on the fairness of classifiers $g_{s,t}$ induced by $s$ hold for all
possible thresholds $t$.
While this strong property is in principle desirable, it puts overly
restrictive constraints on $s$ that will often
completely jeopardize its ranking performance.

We thus propose a general approach to implement the satisfaction of a \emph{finite} number
of fairness constraints on $\Delta_{H,\alpha}(s)$ and $\Delta_{G,\alpha}(s)$ for
specific values of
$\alpha$ that are relevant to the use case at hand.
Our criterion 
is flexible enough to address the limitations of $\auc$-based
constraints outlined above. Specifically, a practitioner can
choose points for $\Delta_{H, \alpha}$ and $\Delta_{G, \alpha}$
to guarantee the fairness
of classifiers obtained by thresholding the scoring function at the
desired trade-offs between, say, FPR and FNR. Furthermore,
we show in \cref{prop:discretization-result} below (proof in supplementary)
that under some regularity assumption on the $\roc$ curve (\cref{ass:assumption-roc-reg}),
if a small number of fairness constraints $m_F$ are satisfied at discrete
points $\alpha_F^{(1)},\dots,\alpha_F^{(m_F)}$
of an interval for $F \in \{H,G\}$,
then one obtains guarantees in sup norm on $\alpha \mapsto \Delta_{F, \alpha}$
(and therefore fair classifiers) in the entire interval $[\alpha_F^{
(1)},\alpha_F^{(m_F)}]$.
This result is crucial in applications where the threshold used at deployment
can vary in a whole interval, such as biometric verification \citep{Grother2019}
 and credit-risk screening (see Example~\ref{ex:credit}).

\begin{assumption}
    The class $\mathcal{S}$ of scoring functions take values
    in $(0,T)$ for some $T > 0$, and the family of cdfs
    $\mathcal{K} = \{ G_s^{(z)}, H_s^{(z)}: s \in \mathcal{S}, z \in \{0,1\} \}$ 
    satisfies: (a) any $K\in\mathcal{K}$ is continuously differentiable, 
    and (b) there exists $b, B > 0$ s.t. $\forall (K, t) \in \mathcal{K}
    \times (0,T)$,
    $b \le \abs{K'(t)} \le B$.
    The latter condition is satisfied when scoring functions
do not have flat or steep parts, see
\cite{Clemencon2007} (Remark 7) for a discussion.
    \label{ass:assumption-roc-reg}
\end{assumption}

\begin{proposition}\label{prop:discretization-result}
    Under \cref{ass:assumption-roc-reg}, if $\exists F \in
    \{H,G\}$ s.t.
    for every $k \in \{1, \dots, m_F\}$,
    $|\Delta_{F, \alpha_F^{(k)}}(s)| \le \epsilon$, then:
    \begin{align*}
    \sup_{\alpha \in [0,1]} \big|\Delta_{F, \alpha}(s)\big|
    \le \epsilon +
    \frac{B+b}{2b}\max_{k\in \{0, \dots, m\}}
    \big|\alpha_F^{(k+1)} - \alpha_F^{(k)}\big| ,
    \end{align*} 
    with the convention $\alpha_F^{(0)} = 0$ and $\alpha_F^{(m_F+1)} = 1$.
\end{proposition}

To illustrate how $\roc$-based fairness constraints can be designed
in a practical case, we return to our credit lending example. In
\cref{fig:example-1} (bottom right), 
 we have
$\Delta_{H, \alpha}(s) = 0$ for any $\alpha \in [0,1]$ since 
$H_s^{(0)} = H_s^{(1)}$.
However, $\Delta_{G, \alpha}(s)$ can be large: this is the case in particular 
for small $\alpha$'s (low FPR).
If the goal is to obtain fair classifiers in FNR
for high thresholds (i.e., low FPR),
we should seek a scoring function $s$ with $\Delta_{G, \alpha}
\simeq 0$ for any $\alpha \le \alpha_{\text{max}}$,
where $\alpha_{\text{max}}$ is the maximum TPR the bank will operate at 
(see \cref{fig:example-1}, bottom right).
The value of $\alpha_{\text{max}}$ can be chosen based on the performance of a
score learned without fairness constraint if the bank seeks to limit FPR's or
maximize its potential earnings. Learning with 
constraints for $\alpha$'s in an evenly
spaced grid on $[0, \alpha_{\text{max}}]$ will ensure that the resulting $s$
yields fair classifiers $g_{s,t}$ for high thresholds $t$, as confirmed
experimentally in Section~\ref{sec:experiments}.

\section{LEARNING UNDER AUC AND ROC FAIRNESS CONSTRAINTS}\label{sec:beyond_auc_cons}

In this section, we first introduce empirical risk minimization problems
for learning
under the $\auc$ and $\roc$-based constraints introduced in Section~\ref{sec:fairness_ranking}.
Then, we prove statistical learning guarantees in the form of 
generalization bounds,
which fill a gap in the existing literature for $\auc$-based constraints
and provide a theoretical justification for our novel $\roc$-based
constraints.
Finally, we briefly describe how to empirically minimize such criteria with
gradient-based algorithms.

\subsection{Learning with AUC-based Constraints}
\label{subsec:learning-under-auc-cons}

We first formulate the problem of bipartite ranking under $\auc$-based
fairness
constraints.
Introducing fairness as a hard constraint is tempting, but may be costly
in terms of ranking performance.
In general,
there is indeed a trade-off between the ranking performance and the level
of fairness.
For a family of scoring
functions $\mathcal{S}$ and some instantiation
$\Gamma$ of our general fairness definition in 
\cref{eq:barycenter-constraint-formulation}, we thus define the learning
problem as follows:
\begin{align}
\label{eq:auc_general_problem}
    \textstyle\max_{s\in\mathcal{S}} \quad \auc_{H_s,G_s} - \lambda 
    \abs{\Gamma^\top
    C
    (s)},
\end{align}
where $\lambda\ge 0$ is a hyperparameter balancing ranking performance
and fairness.

For the sake of simplicity and concreteness, in the rest of this section we
focus on a special case of \cref{eq:auc_general_problem}, namely
when $C(s)$ corresponds to the fairness definition 
in \cref{rk-cons:intra-pairwise}.
One can easily extend our analysis to any other instance of our general
definition
in \cref{eq:barycenter-constraint-formulation}.
We denote by $s_\lambda^*$ the scoring function that maximizes the
objective $L_\lambda(s)$ of \cref{eq:auc_general_problem}, where:
\begin{align*}
    L_\lambda(s) :=
    \auc_{H_s,G_s} - \lambda \big|\auc_{H^{(0)}_s, G^{(0)}_s} - \auc_{ H^{
    (1)}_s,
G^{(1)}_s }\big|.
\end{align*}

Given a training set $\{
(X_i, Y_i, Z_i) \}_{i=1}^n$ of $n$ i.i.d. copies of the random triplet $
(X,Y,Z)$, we denote by $n^{(z)}$ the
number of points in group $z\in\{0,1\}$, and by $n_{+}^{(z)}$ (resp. $n_-^{
(z)}$) the
number of positive (resp. negative) points in $z$. The empirical
counterparts of $H^{(z)}_s$ and $G^{(z)}_s$ are:
\begin{align*}
    \widehat{H}_s^{(z)} (t) & = (1/n_-^{(z)})
    \textstyle\sum_{i=1}^n \I \left\{ Z_i = z, Y_i = - 1, s(X_i) \le t
    \right\}, \\
    \widehat{G}_s^{(z)} (t) & = (1/n_+^{(z)})
    \textstyle\sum_{i=1}^n \I \left\{ Z_i = z, Y_i = +1, s(X_i) \le t \right\}.
\end{align*}
Recalling the notation $\widehat{\auc}_{F,F'} := \auc_{\widehat{F}, 
\widehat{F}'}$ from \cref{subsec:bipartite-ranking},
the empirical problem writes:
\begin{align*}
    \widehat{L}_\lambda(s) :=
    \widehat{\auc}_{H_s,G_s} 
    &- \lambda \big|\widehat{\auc}_{H^{(0)}_s, G^{(0)}_s} 
    - \widehat{\auc}_{H^{(1)}_s, G^{(1)}_s} \big|.
\end{align*}
We denote its maximizer by $\widehat{s}_\lambda$.
We can now state our statistical learning guarantees for fair
ranking.

\begin{theorem}\label{th:gen_auc_cons}
    Assume the class of functions $\mathcal{S}$ is VC-major with finite
	    VC-dimension $V < +\infty$ and that there exists $\epsilon > 0$ s.t.
		$\min_{ z \in \{0,1\},y\in\{-1,1\}} \p\{ Y= y , Z=z \}
                \ge \epsilon$.
    Then, for any $\delta > 0$, for all $n > 1$,
    we have w.p. at least $1-\delta$:
    \begin{align*}
	  \epsilon^2 &\cdot \left[
		  L_\lambda(s_\lambda^*) 
		  - L_\lambda(\widehat{s}_\lambda) 
	      \right]  \le \;  C\sqrt{V/n} \cdot
	 \left( 4\lambda + 1/2 \right)\\
	 & \qquad  
	 + \sqrt{\frac{\log(13/\delta)}{n-1}} \cdot
	 \Big( 4\lambda  + (4\lambda + 2) \epsilon \Big )
	 + O(n^{-1}).
    \end{align*}
\end{theorem}

Theorem~\ref{th:gen_auc_cons} establishes a learning rate of $O(1/
\sqrt{n})$ for our problem of ranking under AUC-based fairness
constraints, which holds for any distribution of $(X,Y,Z)$ as long
as the probability of observing each combination of label and group is bounded
away from zero.
As the natural estimate of the $\auc$ involves sums of dependent
random variables, the proof of \cref{th:gen_auc_cons} does not follow from usual
concentration inequalities on standard averages.
Indeed, it requires controlling the uniform deviation of ratios of $U$-processes indexed
by a class of functions of controlled complexity.

\subsection{Learning with ROC-based Constraints}
\label{sec:sec4-statistical-guarantees}

We now turn to the problem of bipartite ranking under $\roc$-based fairness
constraints. Recall from Section~\ref{sec:new-pointwise-roc} that we aim to
satisfy some constraints on $\Delta_{H,\alpha}(s)$ and $\Delta_{G,\alpha}(s)$
for
specific values of
$\alpha$.
Denote by $m_H,m_G \in \N$ be the number of constraints
for the negatives and the positives respectively, as well as $\alpha_{H} = 
[\alpha_H^{(1)},\dots,\alpha_H^{(m_H)}]\in[0,1]^{m_H}$ and $\alpha_{G} = 
[\alpha_G^{(1)},\dots,\alpha_G^{(m_G)}]\in[0,1]^{m_G}$ the points at which they
apply 
(sorted in strictly increasing order).

With the notation $\Lambda:=(\alpha, \lambda_H, \lambda_G)$,
we can introduce the learning objective $L_\Lambda(s)$ defined as:
\begin{align*}
    \auc_{H_s,G_s} &- 
    \sum_{k=1}^{m_H} \lambda_H^{(k)}  \big| \Delta_{H,\alpha_H^{
    (k)}}(s) \big| 
    - \sum_{k=1}^{m_G} \lambda_G^{(k)} \big| \Delta_{G,\alpha_G^{(k)}}(s) \big|,
\end{align*}
where 
$\lambda_H=[\lambda_H^{(1)},\dots,\lambda_H^{(m_H)}]\in \R_+^
{m_H}$ and
$\lambda_G=[\lambda_G^{(1)},\dots,\lambda_G^{(m_G)}]\in \R_+^{m_G}$ are
hyperparameters.

The empirical counterpart $ \widehat{L}_\Lambda(s)$ of
$L_\Lambda$ is defined as:
\begin{align*}
    \widehat{\auc}_{H_s,G_s} - 
    \sum_{k=1}^{m_H} \lambda_H^{(k)}  \big| \widehat{\Delta}_
    {H,\alpha_H^{(k)}}(s) \big|
    - \sum_{k=1}^{m_G} \lambda_G^{(k)}
    \big| \widehat{\Delta}_{G,\alpha_G^{(k)}}(s) \big|,
\end{align*}
where $\widehat{\Delta}_{H,\alpha}(s)= \widehat{\roc}_{G^{(0)}_s, G^{(1)}_s}
    (\alpha) - \alpha$ and $\widehat{\Delta}_{G,\alpha}(s)=\widehat{\roc}_{H^{(0)}_s,H^{(1)}_s}
    (\alpha) - \alpha$
for any $\alpha \in [0, 1]$.

We now prove statistical guarantees for the maximization of
$ \widehat{L}_\Lambda(s)$.
We denote by $s_\Lambda^*$ the maximizer of $L_\Lambda$ over
$\mathcal{S}$, and by $\hat{s}_\Lambda$ the maximizer of $\widehat{L}_\Lambda$
over $\mathcal{S}$.
Our analysis relies on the regularity assumption on the $\roc$
curve provided in \cref{sec:new-pointwise-roc} (\cref{ass:assumption-roc-reg}).

\begin{theorem}\label{th:gen_fair_roc}
    Under  \cref{ass:assumption-roc-reg} and those of \cref{th:gen_auc_cons},
    for any $\delta > 0$, $n > 1$, w.p. $\ge 1-\delta$:
    \begin{align*}
     & \epsilon^2 \cdot 
    \left[ L_\Lambda(s_\Lambda^*) - L_\Lambda(\widehat{s}_\Lambda) \right]  
    \le  
    C \left( 1/2 + 2 \epsilon C_{\Lambda, \mathcal{K}} \right)
    \sqrt{ V/n}\\
    & \qquad  \qquad \qquad
    +  
    2\epsilon\left( 1 + 3 C_{\Lambda, \mathcal{K}} \right) 
    \sqrt{ \frac{\log(19/\delta)}{n-1} }   
    + O(n^{-1}),
    \end{align*}
    where $C_{\Lambda, \mathcal{K}}= (1+ B/b)(\bar{\lambda}_H + \bar{\lambda}_G)$,
    with
    $\bar{\lambda}_H = \sum_{k=1}^{m_H} \lambda_H^{(k)}$
    and $\bar{\lambda}_G = \sum_{k=1}^{m_G} \lambda_G^{(k)}$.
\end{theorem}

\cref{th:gen_fair_roc} generalizes the learning rate of $O(1/\sqrt{n})$
of \cref{th:gen_auc_cons} to ranking under $\roc$-based constraints.
Its proof also relies on results for
$U$-processes, but further requires a study of the deviations of
the empirical $\roc$ curve
seen as ratios of empirical processes indexed by $\mathcal{S} \times [0, 1]$.
In that regard, our analysis builds upon the decomposition
proposed in \cite{Hsieh1996}, which enables the derivation of uniform
bounds over $\mathcal{S} \times [0, 1]$ from results on standard empirical
processes \citep{VanderVaart1996}.

\subsection{Algorithmic Details}

In practice, maximizing $\widehat{L}_\lambda$ 
or $\widehat{L}_\Lambda$ 
directly by gradient ascent is
not feasible since the criteria are not continuous.
We use classic smooth surrogate relaxations of the AUCs
or $\roc$s based on the logistic function $\sigma:x \mapsto 1/(1+e^{-x})$.
We also remove the absolute values in $\widehat{L}_\lambda$ and $
\widehat{L}_\Lambda$, and instead rely on parameters that are
modified adaptively during the training process.
We solve the problem using a stochastic gradient ascent algorithm,
and modify the introduced parameters every fixed number of iterations
based on fairness statistics evaluated on a small validation set. We refer to
the supplementary material for more details on the algorithms we use in our
experiments.

\begin{figure}[t]
\centering
\includegraphics[width=0.73\columnwidth]{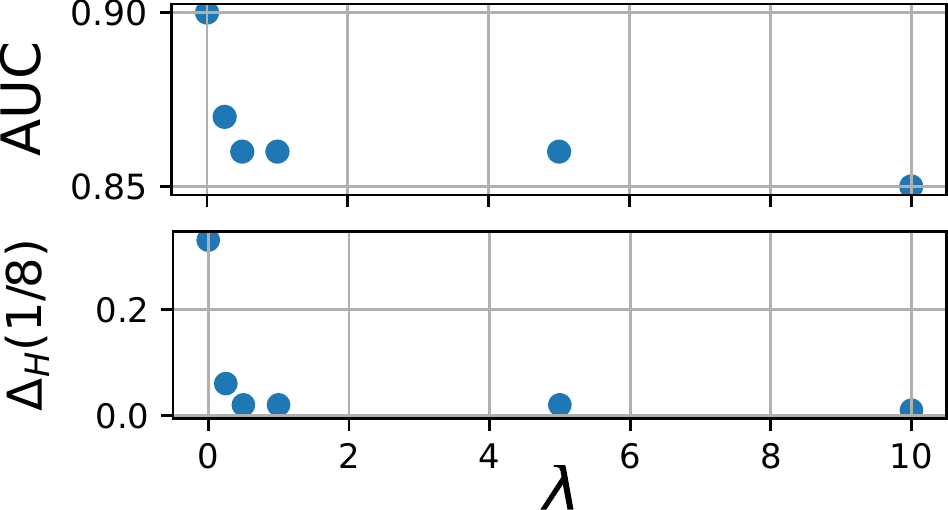}
\caption{Ranking accuracy ($\auc$) and a $\roc$-based constraint 
at $\Delta_H(1/8)$ as a function of the hyperparameter $\lambda$, on the Adult
dataset.
}\label{fig:tradeoff_lambda}
\end{figure}

The hyperparameter $\lambda$ should be tuned to achieve the desired trade-off
between ranking performance and fairness.
For learning under a $\roc$-based constraint,
\cref{fig:tradeoff_lambda} provides 
examples of trade-offs for different $\lambda$'s
on the dataset \textit{Adult} presented in \cref{sec:experiments}.

\section{EXPERIMENTS}\label{sec:experiments}

In this section, 
we present a subset of our experimental results,
which we think nicely illustrates the differences between $\auc$ and
$\roc$-based fairness. It also shows how these constraints can be used to
achieve a trade-off between ranking performance
and the desired notion of fairness in practical use cases. 
Due to space limitations, we refer to the supplementary material for the
presentation of all details on the experimental setup, as
well as
additional results.

\begin{figure*}
\centering
\includegraphics[width=\textwidth]{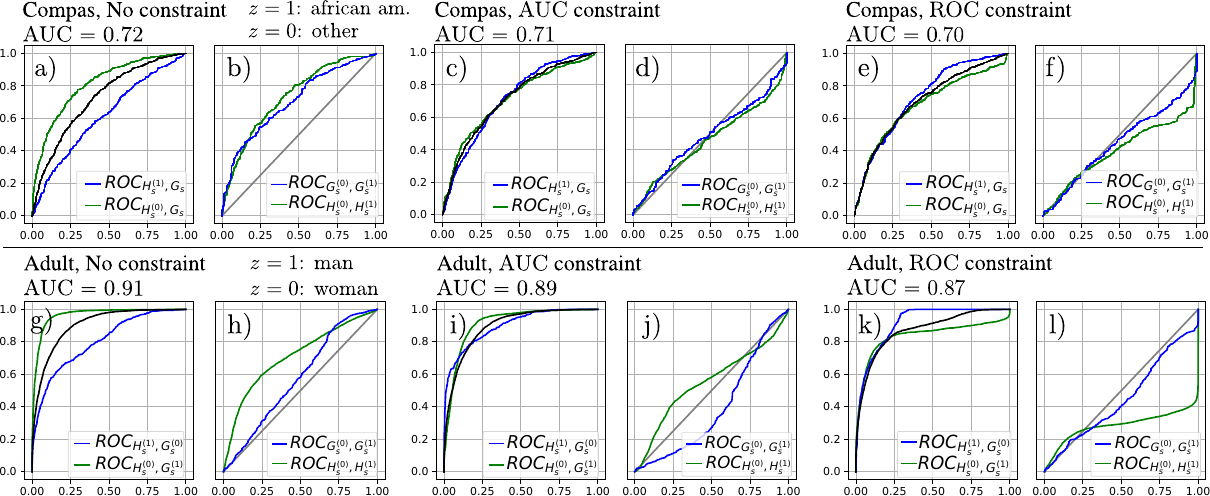}
    \caption{$\roc$ curves on the test set of Adult and Compas
	for a score learned without and with fairness constraints.
	Black curves represent $\roc_{H_s,G_s}$. We also report the corresponding
   ranking performance $\auc_{H_s,G_s}$.}
    \label{fig:roc_real_main_text}
\end{figure*}
Results are summarized in \cref{fig:roc_real_main_text}, which shows
$\roc$ curves for 2-layer
neural scoring functions learned with and without fairness constraints on 2
real
datasets: \textit{Compas} and \textit{Adult} \citep[used \emph{e.g.} in][]{Donini2018}.

\textit{Compas} is a recidivism prediction dataset. We define the sensitive
variable to be $Z = 1$ if the individual is categorized as African-American
and $0$ otherwise.
In contrast to credit-risk screening, here being labeled positive (i.e.,
recidivist) is a disadvantage, so
we consider the \textit{Background Positive Subgroup Negative (BPSN)
$\auc$ fairness constraint} defined as $\auc_{ H^{(0)}_s, G_s } = \auc_{ H^{
(1)}_s, G_s }$, which is equivalent to \cref{rk-cons:pairwise-accuracy}
with positive and negative labels swapped.
BPSN forces the probabilities
that a negative from a given group is mistakenly ranked higher
than a positive to be the same across groups.
While the scoring function learned without fairness constraint
systematically makes more ranking errors for non-recidivist
African-Americans (Fig. \ref{fig:roc_real_main_text}-a),
we can see that learning with the $\auc$-constraint achieves its goal as
it makes the area
under $\roc_{H_s^{(1)},G_s}$ and $\roc_{H_s^{(0)},G_s}$ very similar
(Fig. \ref{fig:roc_real_main_text}-c).
However, slightly more of such errors are still made in the
top 25\% of the scores, which is the region where the prediction threshold
could be set in practice for taking decisions such as denying bail.
We thus configure our ROC-based
fairness constraints to align the
distributions of positives and negatives across both groups by penalizing
solutions with high $|\Delta_{G, 1/8}(s)|$, $|\Delta_{G, 1/4}(s)|$, $|\Delta_
{H, 1/8}(s)|$ and $|\Delta_{H, 1/4}(s)|$. In line with our theoretical
analysis (see the discussion in Section~\ref{sec:new-pointwise-roc}), we can
see from $\roc_{G_s^{(0)},G_s^{(1)}}$ and $\roc_{H_s^{(0)},H_s^{(1)}}$ that
this suffices to learn a scoring function that achieves equality of the
positive and negative distributions in the entire interval $[0,1/4]$ of
interest (Fig. \ref{fig:roc_real_main_text}-f).
In turn, $\roc_{H_s^{(1)},G_s}$ and $\roc_{H_s^{(0)},G_s}$ become essentially
equal in this region as desired (Fig. \ref{fig:roc_real_main_text}-e).
Note that on this dataset, both the $\auc$ and $\roc$
constraints are achieved with minor impact on the ranking performance, as seen
from the AUC scores.

We now turn to the \textit{Adult} dataset, where we set $Z$
to denote the gender ($0$ for female) and $Y=1$ indicates that
the person
makes over \$$50$K/year.
For this dataset, we plot $\roc_{H_s^{(1)},G_s^{(0)}}$ and $\roc_
{H_s^{(0)},G_s^{(1)}}$ and observe that without fairness constraint, men who
make less than \$$50$K are much more likely to be mistakenly ranked above a
woman who actually makes more, than the other way around
(Fig. \ref{fig:roc_real_main_text}-g). 
The learned score thus reproduces a common gender bias.
To fix this, the appropriate notion of $\auc$-based fairness is 
\cref{rk-cons:inter-group-pairwise-fairness}. We see that learning under this
constraint successfully
equates the area under $\roc_{H_s^{(1)},G_s^{(0)}}$ and $\roc_
{H_s^{(0)},G_s^{(1)}}$ (Fig. \ref{fig:roc_real_main_text}-i). However,
this comes at the cost of introducing a small
bias against men in the top scores. As seen from $\roc_{H_s^{(0)},H_s^{
(1)}}$ and $\roc_{G_s^{(0)},G_s^{(1)}}$, positive women now have higher
scores overall than positive men, while negative men have higher
scores than negative women (Fig. \ref{fig:roc_real_main_text}-j). These
observations illustrate the limitations of
$\auc$ fairness (see Section~\ref{sec:limits-auc-based-constraints}). To
address them, we use the same $\roc$ constraints as for \emph{Compas}
so as to align the positive and negative distributions of each
group in $[0, 1/4]$. This is again achieved almost perfectly in the
entire interval (Fig. \ref{fig:roc_real_main_text}-l). While the degradation in
ranking performance is more
noticeable on this dataset, a clear advantage from $\roc$-based fairness is
that the scoring function can be thresholded to
obtain fair classifiers at a wide range of thresholds.

\section{DISCUSSION}\label{sec:conclusion}

In this work, we studied the problem of fairness for scoring functions
learned from binary labeled data. 
We proposed a general framework for designing $\auc$-based fairness
constraints, introduced novel $\roc$-based constraints, and derived
statistical guarantees for learning scoring functions under such
constraints.
Although we focused on $\roc$ curves, our framework can be
adapted to
\emph{precision-recall curves} (as they are a function of the FPR and TPR 
\citep{PR_curve}). It can also be extended to \textit{similarity ranking}, a
variant of
bipartite
ranking covering applications
like biometric authentication
\citep{pmlr-v80-vogel18a}.

Recent work derived analytical expressions of optimal fair models
for learning problems other than bipartite ranking 
\citep{Menon18,FairRegPaper}.
A promising direction for future work is to derive a similar result
for scoring functions. This would enable us to
propose a compelling
theoretical study of the trade-offs between performance and fairness
in bipartite ranking, and lay the foundations 
for provably fair extensions of 
$\roc$ curve optimization algorithms based on recursive partitioning \citep{CDV09,CV09CA}.

\bibliography{main}
\bibliographystyle{abbrvnat}

\onecolumn

\section*{SUPPLEMENTARY MATERIAL}
\appendix


\section{Generality of our Family of AUC-Based Fairness Definitions}

In this section, we 
show that 
the framework for AUC-based fairness
we introduced in Section~\ref{subsec:general-auc-fairness-formulation}.
can recover AUC-based fairness constraints introduced in previous work.
Then, we give examples of new fairness constraints that can be expressed with our framework.


\textbf{Recovering existing AUC-based fairness constraints.}
We first expand on the $\auc$-based fairness constraints introduced briefly in
\cref{sec:preliminaries} of the main text, \textit{i.e.}
\cref{rk-cons:intra-pairwise},
\cref{rk-cons:pairwise-accuracy} and
\cref{rk-cons:inter-group-pairwise-fairness}.
First, note that the \textit{intra-group pairwise $\auc$ fairness}
(\cref{rk-cons:intra-pairwise})
is also featured in \cite{Borkan2019} under the name of \textit{subgroup $\auc$ fairness}.
It requires the ranking performance to be equal \emph{within} groups,
which is relevant for instance in applications where groups are ranked separately
(\emph{e.g.}, candidates for two types of jobs).
\cref{rk-cons:pairwise-accuracy} is also featured in \cite{Beutel2019FRR}
under the name of \emph{pairwise accuracy}.
It can be seen as the ranking counterpart of
\emph{parity in false negative rates} in binary classification \cite{Hardt16}.
Finally, \cref{rk-cons:inter-group-pairwise-fairness} is also featured in \cite{Beutel2019FRR}
under the name of \emph{$\text{x}\auc$ parity}.

Other $\auc$-based fairness constraints were introduced in previous work.
Precisely, the constraint used in \cref{sec:experiments} for the 
\textit{Compas}
dataset is featured in \cite{Borkan2019,NIPS2019_8604}, and writes:
\begin{align}\label{rk-cons:BPSN-AUC} 
    \auc_{H_s^{(0)}, G_s} &= \auc_{H_s^{(1)}, G_s}.
\end{align}
\cite{Borkan2019} refers to \cref{rk-cons:BPSN-AUC}
as \emph{Backgroup Positive Subgroup Negative (BPSN)} AUC fairness,
which can be seen as the ranking counterpart of
\emph{parity in false positive rates} in classification \cite{Hardt16}.

Both \cite{Borkan2019} and \cite{NIPS2019_8604} also introduce the following
$\auc$ fairness constraint:
\begin{align}\label{rk-cons:average-equality-gap} 
    \auc_{G_s, G^{(0)}_s} &= \auc_{G_s, G^{(1)}_s}.
\end{align}
\cite{Borkan2019} also defines the \emph{Average Equality Gap (AEG)}
as $\auc(G_s, G_s^{(z)})-1/2$ for $z\in\{0,1\}$.
\cref{rk-cons:average-equality-gap}
thus corresponds to an AEG of zero, 
\textit{i.e.} the scores of the
positives of any group are not stochastically larger than those of the other.

All these AUC-based fairness constraints can be written as
instances of our general definition for a specific choice of $\Gamma$,
as presented in \cref{tab:table_lambdas}.
Note that $\Gamma$ might depend on the quantities $q_0, p_0, q_1, p_1$.

\textbf{Expressing new AUC-based fairness constraints.}
Relevant fairness constraints that have not been considered in previous work
can be expressed using our general
formulation. Denoting $F^{(0)}_s = (1-p_0) H^{(0)}_s + p_0 G^{
(0)}_s$, consider for instance the following constraint:
\begin{align}\label{rk-cons:new-with-F} 
    \auc_{F^{(0)}_s, G^{(0)}_s} = \auc_{F^{(0)}_s, G^{(1)}_s}.
\end{align}
It equalizes the expected position of the positives of each group with
respect to a \emph{reference group} (here group $0$).
Another fairness constraint of interest is based on the rate of misranked
pairs when one element is in a specific group:
\begin{align*}
    E(s, z) := 
    \p  \{  (s(X) - s(X))(Y-Y') > 0 \mid Y \ne Y', Z=z \} + 
    \frac{1}{2} \cdot \p\{ s(X) = s(X) \mid Y \ne Y', Z=z \}.
\end{align*}
The equality $E(s, 0) = E(s, 1)$ can be seen as the analogue of 
\emph{parity in mistreatment} for the task of ordering pairs, see
\cref{eq:parity-in-mistreatment}. It is easy to see that this constraint can
be written in the form of \cref{eq:fairness-mixture} and that point 1 of 
\cref{th:generality_fairness_constraint} holds, hence it is equivalent to
$\mathcal{C}_\Gamma(s)$ for some $\Gamma \in \R^5$.

\section{Relations Between Fairness in Bipartite Ranking and Fairness in
Classification}

In this section, we clarify the relationship between known propositions
for fairness in classification on the one hand, and our $\auc$-based and
$\roc$-fairness for bipartite ranking on the other hand. In a nutshell, we
show that: (i) if a scoring function $s$ satisfies an $\auc$-based fairness
constraint, there exists a certain threshold $t$ such that the classifier $g_
{s,t}$ obtained by thresholding $s$ at $t$ satisfies a fair
classification constraint, and (ii)
$\roc$-based fairness constraints allow to directly control the value of $t$
for which $g_{s,t}$ is fair, and more generally to achieve classification
fairness for a whole range of thresholds, which is useful to address
task-specific operational constraints such as those described in Example~\ref{ex:credit}.

\begin{table}[t]
    \centering
    \caption{Value of $\Gamma = (\Gamma_l)_{l=1}^5$
    for all of the AUC-based fairness constraints
    in the paper for the general formulation of \cref{eq:barycenter-constraint-formulation}.}\label{tab:table_lambdas}
    \vskip 0.15in
    {
    \renewcommand{\arraystretch}{1.25}
    {\footnotesize
    \begin{tabular}[h]{lccccc}
    \hline
    Eq. & $\Gamma_1$ & $\Gamma_2$ & $\Gamma_3$ & $\Gamma_4$ & $\Gamma_5$ \\
    \hline
    (\ref{rk-cons:intra-pairwise}) & $0$ &  $0$ &  $\frac{1}{3}$ & 
    $\frac{1}{3}$ &  $\frac{1}{3}$ \\
    (\ref{rk-cons:pairwise-accuracy}) & $0$ & $0$ & $\frac{q_0(1-p_0)}{1-p}$ & 
    $0$ & $\frac{q_1(1-p_1)}{1-p}$ \\
    (\ref{rk-cons:BPSN-AUC}) & $0$ & $0$ & $\frac{q_0p_0}{2p}$ 
    & $\frac{1}{2}$ & $\frac{q_1p_1}{2p}$ \\
    (\ref{rk-cons:average-equality-gap})  & $0$ & $1$ & $0$ & $0$ & $0$ \\
    (\ref{rk-cons:inter-group-pairwise-fairness}) & $0$ & $0$ & $0$ 
    & $1$ & $0$ \\
    (\ref{rk-cons:new-with-F}) & $0$ & $p_0$ & $1-p_0$ 
    & $0$ & $0$ \\
    \hline
    \end{tabular}
    }
    }
\end{table}

\textbf{Pointwise $\roc$ equality and fairness in binary classification.}
As mentioned in the main text, a scoring function $s: \X \to \R$ induces an
infinite
family of binary classifiers $g_{s, t} : x \mapsto 2\cdot\I\{ s(x) > t \} - 1$
indexed by thresholds $t \in \R$. The following proposition shows that one of
those
classifiers satisfies a fairness constraint as soon
as appropriate group-wise $\roc$ curves are equal for
some value $\alpha \in [0,1]$.
It is proven in \cref{sec:proof-fairness-constraints}.

\begin{proposition}\label{cor:fairness-auc-to-classif}
    Under appropriate conditions on the score function $s$ (\textit{i.e.}, $s\in
    \mathcal{S}$
    where $\mathcal{S}$ satisfies \cref{ass:assumption-roc-reg}),
    we have that:
    \vspace{-\topsep}
    \begin{itemize}\setlength\itemsep{.0em}
	\item If $p_0=p_1$ and $s$ satisfies
	    \begin{align}\label{eq:roc-eq1}
	    \roc_{H_s^{(0)}, G_s^{(0)}} (\alpha)
	    = \roc_{H_s^{(1)}, G_s^{(1)}} (\alpha)
	\end{align}
        for some $\alpha \in [0,1]$, then there exists $(t_0, t_1) \in (0,
        T)^2$,
	s.t.  $M^{(0)}(g_{s,t_0}) = M^{(1)}(g_{s,t_1})$, 
        which resembles parity in mistreatment (see Eq. \ref{eq:parity-in-mistreatment}).
    \item If $s$ satisfies
	\begin{align}\label{eq:roc-eq2}
	    \roc_{H_s, G_s^{(0)}} (\alpha) = \roc_{H_s, G_s^{(1)}} (\alpha)
	\end{align}
        for some $\alpha \in [0,1]$, then $g_{s,t}$ satisfies fairness in FNR
	(see \cref{eq:parity-in-fpr-fnr})
        for some threshold $t \in (0, T)$.
    \item If $s$ satisfies
	\begin{align}\label{eq:roc-eq3}
	    \roc_{H_s^{(0)}, G_s}(\alpha) = \roc_{H_s^{(1)}, G_s}(\alpha)
	\end{align}
	for some $\alpha \in [0,1]$, then $g_{s,t}$ satisfies parity in FPR
    (see \cref{eq:parity-in-fpr-fnr})
        for some threshold $t \in (0, T)$.
    \end{itemize}
\end{proposition}

\textbf{Relation with AUC-based fairness.} For continuous $\roc$s, the
equality between their two $\auc$s implies
that the two $\roc$s intersect at some unknown point, 
as shown by \cref{prop:implications-of-fairness} (a simple consequence of the
mean value theorem) which proof is detailed in \cref{sec:proof-fairness-constraints}.
Theorem 3.3 in
\cite{Borkan2019} corresponds to the special case of
\cref{prop:implications-of-fairness} when $h=g, h' \ne g'$.

\begin{proposition}\label{prop:implications-of-fairness}
    Let $h,g,h',g'$ be cdfs on $\R$ such that $\roc_{h, g}$ 
    and $\roc_{h', g'}$ are continuous.
    If 
    $\auc_{h, g} = \auc_{h',g'}$,
    then there exists $\alpha \in (0,1)$, 
    such that $\roc_{h, g}(\alpha) = \roc_{h', g'}(\alpha)$.
\end{proposition}

\cref{prop:implications-of-fairness}, combined with 
\cref{cor:fairness-auc-to-classif}, implies that
when a scoring function $s$ satisfies
some $\auc$-based fairness constraint, there exists a threshold $t
\in \R$ inducing a non-trivial classifier $g_{s,t}:= \sign(s(x) - t)$
that satisfies some notion of fairness for classification at some unknown
threshold $t$.
For example, it is straightforward from \cref{cor:fairness-auc-to-classif}
and \cref{prop:implications-of-fairness} that:
\begin{itemize}
    \item \cref{rk-cons:intra-pairwise} implies parity in mistreatment for some threshold,
    \item \cref{rk-cons:pairwise-accuracy}, \cref{rk-cons:average-equality-gap}
	and \cref{rk-cons:new-with-F} all imply parity in FNR for some threshold,
    \item \cref{rk-cons:BPSN-AUC} implies parity in FPR for some threshold.
\end{itemize}
The principal drawback of $\auc$-based fairness constraints is that it
guarantees the existence of a single (unknown) $t$ for which the fair binary
classification properties are verified by $g_{s, t}$, and that the
corresponding $\roc$ point $\alpha$ cannot be easily controlled.

\textbf{Relation with ROC-based fairness.}
In contrast to $\auc$-based fairness, $\roc$-based fairness allows to
directly control the points $\alpha$ in \cref{cor:fairness-auc-to-classif} at
which one obtains fair classifiers as it precisely consists in enforcing
equality of $\roc_{G^{(0)}_s, G^{(1)}_s}$ and $\roc_{H^{(0)}_s,H^{
(1)}_s}$ at specific points.

Furthermore, one can impose the equalities
\cref{eq:roc-eq1}, \cref{eq:roc-eq2} and \cref{eq:roc-eq3} for several
values of $\alpha$ such that thresholding the score behaves well for many
critical situations. Specifically, under \cref{ass:assumption-roc-reg},
we prove in \cref{prop:discretization-result} 
of \cref{sec:new-pointwise-roc} (see \cref{sec:proof-fairness-constraints} for the proof)
that pointwise constraints over a discretization of the interval of interest
approximate its satisfaction on the whole interval.
This behavior, confirmed by our empirical results (see
Sections~\ref{sec:experiments} and
\ref{sec:appendix:real_data_experiments}), is relevant for many real-world
problems that requires fairness in
binary classification to be satisfied for a whole
range of thresholds $t$ in a specific region, see the credit risk-screening
use case of Example~\ref{ex:credit}. We can also mention the example of
biometric verification, where one is interested 
in low false positive rates (\textit{i.e.}, large thresholds $t$).
We refer to \cite{Grother2019} for an evaluation of
the fairness of facial recognition systems in the context of 1:1 verification.

\section{Proofs of Fairness Constraints Properties}\label{sec:proof-fairness-constraints}

\subsection{Proof of \cref{th:generality_fairness_constraint}}

Denote $ D(s) = ( D_1(s), D_2(s), D_3(s), D_4(s) )^\top := (H^{(0)}_s, H^{(1)}_s,
G^{(0)}_s, G^{(1)}_s)^\top$.
For any $(i,j)\in\{1,\dots,4\}^2$, we introduce the notation:
\begin{align*}
    \auc_{D_i, D_j} : s \mapsto \auc_{D_i(s), D_j(s)}.
\end{align*}
Introduce a function $M$ such that $M(s) \in \R^{4\times 4}$
for any $s : X \to \R$, and for any $(i,j)\in\{1, \dots, 4\}$,
the $(i,j)$ coordinate of $M$ writes:
\begin{align*}
    M_{i,j} = \auc_{D_i,D_j} - \frac{1}{2}.
\end{align*}
First note that, for any $s$, $M(s)$ is antisymmetric
\textit{i.e.} $M_{j,i}(s) = - M_{i,j}(s)$ for any $(i,j) \in \{1, \dots, 4\}^2$.
Then, with $(\alpha, \beta) \in \mathcal{P}^2$, we have that:
\begin{align*}
    \auc_{\alpha^\top D , \beta^\top D}
    = \alpha^\top M \beta - \frac{1}{2} 
    = \langle M, \alpha \beta^\top \rangle - \frac{1}{2},
\end{align*}
where $\langle M, M' \rangle = \text{tr} ( M^\top M' )$ is the standard dot product
between matrices. \cref{eq:fairness-mixture} can be written as:
\begin{align}\label{eq:weighted-sum}
    \langle M, \alpha \beta^{\top} - \alpha' \beta^{'\top} \rangle = 0.
\end{align}

\textit{Case of $\alpha = \alpha'$ and $\beta - \beta' = \delta (e_i - e_j)$.}

Consider the specific case where $\alpha = \alpha'$ and $\beta - \beta' =
\delta (e_i - e_j)$ with $i \ne j$ and $\delta \ne 0$, then
\begin{align*}
    \langle M, \alpha(\beta-\beta')^\top \rangle = \delta K^{(\alpha)}_{i,j},
\end{align*}
where:
\begin{align*}
     K^{(\alpha)}_{i,j} & = \langle M, \alpha(e_i - e_j) ^\top \rangle 
     = \sum_{k=1}^{4} \alpha_k \left[ \auc_{D_k, D_i} - \auc_{D_k, D_j} \right], \\
    & = (\alpha_i + \alpha_j) \left[ \frac{1}{2} - \auc_{D_i, D_j} \right] +
    \sum_{k\notin \{ i,j \}} \alpha_k \left[ \auc_{D_k, D_i} - \auc_{D_k, D_j} \right],
\end{align*}
The preceding definition implies that $K^{(\alpha)}_{i,j} = - K^{(\alpha)}_{j,i}$.
Using $\sum_{k=1}^K \alpha_k = 0$, we can express every $K^{(\alpha)}_{i,j}$ as
a linear combinations of the $C_l$'s plus a remainder, precisely:
\begin{align*}
    K^{(\alpha)}_{1,2} &= - \left( \alpha_1 + \alpha_2 \right) C_1 - \alpha_3 (
    C_3 + C_4 ) - \alpha_4 (C_4 + C_5), \\
    K^{(\alpha)}_{1,3} &=  \left( \frac{1}{2} - \auc_{D_1, D_3} \right)  
    + \alpha_2 (-C_1 + C_3 + C_4)
    + \alpha_4 (- C_2 + C_3), \\
    K^{(\alpha)}_{1,4} &=  \left( \frac{1}{2} - \auc_{D_1, D_4} \right)  
    + \alpha_2 (- C_1 + C_4 + C_5)
    + \alpha_3 (C_2 - C_3 - C_4), \\
    K^{(\alpha)}_{2,3} &=  \left( \frac{1}{2} -  \auc_{D_2, D_3} \right)  
    + \alpha_1 (C_1 - C_3 - C_4)
    + \alpha_4 (-C_2 + C_5), \\
    K^{(\alpha)}_{2,4} &=  \left( \frac{1}{2} - \auc_{D_2, D_4} \right)  
    + \alpha_1 (C_1 - C_4 - C_5 )
    + \alpha_3 (C_2 - C_5), \\
    K^{(\alpha)}_{3,4} &= 
    \left( \alpha_3 + \alpha_4 \right)  C_2 + \alpha_1 C_3 + \alpha_2 C_5.
\end{align*}

Hence, it suffices that $\{ i,j \} = \{ 1, 2\}$ or $\{ i,j \} = \{ 3,4 \}$
for \cref{eq:weighted-sum} to be equivalent to $\mathcal{C}_\Gamma$ for some $\Gamma \in \R^5$.

\textit{Case of $\alpha = \alpha'$.}

Any of the $\beta - \beta'$ can be written as a positive linear combination of
$e_i - e_j$ with $i \ne j$, since:
\begin{align*}
    \beta-\beta' 
    = \frac{1}{4} \sum_{i\ne j}  \left( \beta_i + \beta_j' \right) \left( e_i - e_j \right),
\end{align*}
which means that, since $K^{(\alpha)}_{i,j} = - K^{(\alpha)}_{j,i}$:
\begin{align}\label{eq:generality-fairness-dec}
    \langle M, \alpha ( \beta - \beta' )^\top \rangle 
    =  \frac{1}{4} \sum_{i\ne j} \left( \beta_i + \beta_j' \right) K^{(\alpha)}_{i,j} 
    =  \frac{1}{4} \sum_{i<j} \left( [\beta_i - \beta_j]
    - [\beta_i' - \beta_j'] \right) K^{(\alpha)}_{i,j}.
\end{align}
Note that any linear combination of the $K^{(\alpha)}_{1,3}$, $K^{(\alpha)}_{1,4}$,
$K^{(\alpha)}_{2,3}$ and $K^{(\alpha)}_{2,4}$:
\begin{align*}
    \gamma_1 \cdot K^{(\alpha)}_{1,3}
    + \gamma_2 \cdot K^{(\alpha)}_{1,4}
    + \gamma_3 \cdot K^{(\alpha)}_{2,3}
    + \gamma_4 \cdot K^{(\alpha)}_{2,4},
\end{align*}
where $\gamma \in R^4$ with $\mathbf{1}^\top\gamma = 0$
can be written as a weighted sum of the $C_l$ for $l\in\{1,\dots,5\}$.

Hence, it suffices that $\beta_1 + \beta_2 = \beta_1' + \beta_2'$
for \cref{eq:generality-fairness-dec} to be equivalent to some
$\mathcal{C}_\Gamma$ for some $\Gamma \in \R^5$.

\textit{General case.}

Note that, using the antisymmetry of $M$ and \cref{eq:generality-fairness-dec}:
\begin{align*}
    \langle M, \alpha \beta^{\top} - \alpha' \beta^{'\top} \rangle 
    & = \langle M, \alpha (\beta - \beta')^\top \rangle 
    + \langle M, (\alpha - \alpha')\beta'^{\top}  \rangle, \\
    & = \langle M, \alpha (\beta - \beta')^\top \rangle 
    - \langle M, \beta' (\alpha - \alpha')^{\top}  \rangle, \\
    & =  \frac{1}{4} \sum_{i<j} \left[
    \left( [\beta_i - \beta_j] - [\beta_i' - \beta_j'] \right) K^{(\alpha)}_{i,j}
    - \left( [\alpha_i - \alpha_j] - [\alpha_i' - \alpha_j'] \right) K^{(\beta')}_{i,j} 
    \right], 
\end{align*}
Hence, it suffices that $(e_1 + e_2)^\top [ (\alpha - \alpha') - (\beta-\beta')] = 0$
for \cref{eq:weighted-sum} to be equivalent to some
$\mathcal{C}_\Gamma$ for some $\Gamma \in \R^5$.

\textit{Conclusion.}

We denote the three propositions of \cref{th:generality_fairness_constraint}
as $P_1$, $P_2$ and $P_3$.

Assume that $H^{(0)} = H^{(1)}$, $G^{(0)} = G^{(1)}$ and $\mu(\eta(X) = 1/2) < 1$,
then $C_l = 0$ for any $l \in \{1, \dots, 5\}$, which gives:
\begin{align*}
    \langle & M(s), \alpha \beta^{\top} - \alpha' \beta^{'\top} \rangle \\
    & = \frac{1}{4} 
    \left( \frac{1}{2} - \auc_{H_s,G_s} \right)
    \left(
    \sum_{i \in \{1,2\}} \sum_{j\in\{3,4\}} 
    \left[ \left( [\beta_i - \beta_j] - [\beta_i' - \beta_j'] \right) 
    - \left( [\alpha_i - \alpha_j] - [\alpha_i' - \alpha_j'] \right) 
    \right]
    \right),  \\
    & = \left( \frac{1}{2} - \auc_{H_s,G_s} \right)
     (e_1 + e_2)^\top [ (\alpha - \alpha') - (\beta - \beta') ] ,
\end{align*}
It is known that:
\begin{align*}
    \auc_{H_\eta,G_\eta} = \frac{1}{2} 
    + \frac{1}{4 p(1-p)} \iint \abs{\eta(x) - \eta(x')} d \mu(x) d \mu(x'),
\end{align*}
which means that $\auc_{H_\eta,G_\eta} = 1/2$ implies that $\eta(X) = p$ almost surely (a.s.),
and the converse is true.

Assume $P_1$ is true, then $\auc_{H_\eta,G_\eta} > 1/2$, 
hence $(e_1 + e_2)^\top [ (\alpha - \alpha') - (\beta - \beta') ] = 0$
because \cref{eq:weighted-sum} is verified for $\eta$,
and we have shown $P_1 \implies P_3$.

Assume $P_3$ is true, then $\langle M, \alpha \beta^{\top} - \alpha' \beta^{'\top} \rangle$
writes as a linear combination of the $C_l$'s, $l \in \{1, \dots, 5\}$,
and we have shown that $P_3 \implies P_2$.

Assume $P_2$ is true, then observe that if $H^{(0)} = H^{(1)}$ and $G^{(0)} = G^{(1)}$,
then any $\mathcal{C}_\Gamma$ is satisfied for any $\Gamma \in \R^5$, and we have shown 
that $P_2 \implies P_1$, which concludes the proof.

\subsection{Proof of \cref{cor:fairness-auc-to-classif}}

We go over each case.

\textbf{Case of \cref{eq:roc-eq1}.}
\cref{eq:roc-eq1} also writes:
\begin{align*}
    G_s^{(0)} \circ \left( H_s^{(0)}\right)^{-1}(\alpha)
    = G_s^{(1)} \circ \left(H_s^{(1)} \right)^{-1}(\alpha),
\end{align*}
Introduce $t_z= (H^{(z)}_s)^{-1}(\alpha)$ 
then $G^{(z)}_s(t_z) = H^{(z)}_s(t_z) = \alpha$ for any $z \in \{0,1\}$, since
$H_s^{(z)}$ is increasing. Also,
\begin{align*}
    M^{(z)} (g_{s,t_z}) = \p\{ g_{s,t_z}(X) \ne Y \mid Z = z\} 
    = p_z G_s^{(z)}(t_z) + (1-p_z) (1-H_s^{(z)}(t_z))
    = (2\alpha-1)p_z + (1-\alpha),
\end{align*}
which implies the result.

\textbf{Case of \cref{eq:roc-eq2}.}
\cref{eq:roc-eq2} also writes:
\begin{align*}
    G_s^{(0)} \circ H_s^{-1}(\alpha) = G_s^{(1)} \circ H_s^{-1}(\alpha),
\end{align*}
which translates to:
\begin{align*}
    G^{(0)} \left( s(X) \le  H_s^{-1}(\alpha) \right)
    = G^{(1)} \left( s(X) \le  H_s^{-1}(\alpha) \right),
\end{align*}
hence $g_{s,t}$ satisfies fairness in FNR (\cref{eq:parity-in-fpr-fnr})
for the threshold $t = H_s^{-1}(\alpha)$.

\textbf{Case of \cref{eq:roc-eq3}.}
\cref{eq:roc-eq3} also writes:
\begin{align*}
    G_s \circ (H_s^{(0)})^{-1}(\alpha) = G_s \circ (H_s^{(1)})^{-1}(\alpha),
\end{align*}
which implies, since $G_s$, $H_s^{(0)}$ and $H_s^{(1)}$ are increasing:
\begin{align*}
    H_s^{(0)} \circ (H_s^{(0)})^{-1}(\alpha) = H_s^{(1)} \circ (H_s^{(0)})^{-1}(\alpha),
\end{align*}
and:
\begin{align*}
    H^{(0)} \left( s(X) >  (H_s^{(0)})^{-1}(\alpha) \right)
    = H^{(1)} \left( s(X) > (H_s^{(0)})^{-1}(\alpha) \right),
\end{align*}
hence $g_{s,t}$ satisfies fairness in FPR (\cref{eq:parity-in-fpr-fnr})
for the threshold $t = (H_s^{(0)})^{-1}(\alpha)$.

\subsection{Proof of \cref{prop:implications-of-fairness}}

Consider 
$f: [0,1] \mapsto [-1, 1]$:
$f(\alpha) = \roc_{h,g}(\alpha) - \roc_{h',g'}(\alpha)$,
it is continuous, hence integrable, and with:
\begin{align*}
    F(t) = \int_{0}^t f(\alpha) dt,
\end{align*}
Note that $F(1) = \auc_{h,g} - \auc_{h',g'} = 0 = F(0)$.
The mean value theorem implies that there exists $\alpha \in (0,1)$ such that:
\begin{align*}
    \roc_{h,g}(\alpha) = \roc_{h',g'}(\alpha).
\end{align*}

\subsection{Proof of \cref{prop:discretization-result}}

For any $F \in \{H,G\}$, note that:
\begin{align*}
    \sup_{\alpha \in [0,1]} \abs{\Delta_{F, \alpha}(s)}
    \le \max_{k\in \{0, \dots, m\}}
    \sup_{x\in [\alpha_F^{(k)}, \alpha_F^{(k+1)}] }
    \abs{ \Delta_{F, \alpha}(s) }.
\end{align*} 

$\roc_{F_s^{(0)}, F_s^{(1)}}$ is differentiable, and its derivative is
bounded by $B/b$. Indeed, for any $K_1, K_2 \in \mathcal{K}$,
since $K_1$ is continuous and increasing, the inverse function theorem
implies that $(K_1)^{-1}$ is differentiable.
It follows that $K_2 \circ K_1^{-1}$ is differentiable and that
its derivative satisfies:
\begin{align*}
    \left( K_2 \circ K_1^{-1} \right)'
    = \frac{K_2' \circ K_1^{-1}}{K_1' \circ K_1^{-1} }
    \le \frac{B}{b}.
\end{align*}

Let $k \in \{0, \dots, m\}$, and $\alpha \in [\alpha_F^{(k)}, \alpha_F^{(k+1)}]$.
Since $\alpha \mapsto \Delta_{F, \alpha}(s)$ is continuously differentiable,
then $\alpha$ simultaneously satisfies, with the assumption that
$| \Delta_{F, \alpha_F^{(k)}}(s)| \le \epsilon$ for any $k \in \{1, \dots, K\}$:
\begin{align*}
    \abs{ \Delta_{F, \alpha}(s) } \le 
    \epsilon + 
    \left( 1+ \frac{B}{b} \right) \abs{\alpha_F^{(k)} - \alpha} 
    \quad \text{and} \quad
    \abs{ \Delta_{F, \alpha}(s) } \le 
    \epsilon + \left( 1 + \frac{B}{b} \right) \abs{\alpha - \alpha_F^{(k+1)}},
\end{align*}
which implies that
$\abs{ \Delta_{F, \alpha}(s) } \le \epsilon 
+ (1 + B/b) \abs{\alpha_F^{(k+1)} - \alpha_F^{(k)}}/2$.

Finally, we have shown that:
\begin{align*}
    \sup_{\alpha \in [0,1]} \abs{\Delta_{F, \alpha}(s)}
    \le \epsilon +
     \frac{B + b}{2b} \max_{k\in \{0, \dots, m\}}
    \abs{\alpha_F^{(k+1)} - \alpha_F^{(k)}} .
\end{align*} 

\section{Proofs of Generalization Bounds}

\subsection{Definitions}

We recall a few useful definitions.

\begin{definition}[VC-major class of functions -- \citealp{VanderVaart1996}]
\label{def:VC-major}
A class of functions $\mathcal{F}$ such that $\forall f \in \mathcal{F}$, $f:\X \to \R$ is called VC-major
if the major sets of the elements in $\mathcal{F}$ form a VC-class of sets in $\X$.
Formally, $\mathcal{F}$ is a VC-major class if and only if:
\begin{align*}
\left \{  \left \{ x \in \X \mid f(x) > t \right \} \mid f \in \mathcal{F}, t \in \R \right \}
\; \text{is a VC-class of sets.}
\end{align*}
\end{definition}

\begin{definition}[$U$-statistic of degree 2 -- \citealp{Lee1990}]\label{def:U-statistic}
Let $\X$ be some measurable space and $V_1,\; \ldots,\; V_n$ i.i.d. random variables valued in $\X$ and $K:\X^2\rightarrow \mathbb{R}$ a measurable symmetric mapping s.t. $h(V_1,V_2)$ is square integrable. The functional $U_n(h)=(1/n(n-1))\sum_{i\neq j}h(V_i,V_j)$ is referred to as a symmetric $U$-statistic of degree two with kernel $h$. It classically follows from Lehmann-Scheff\'e's lemma that it is the unbiased estimator of the parameter $\mathbb{E}[h(V_1,V_2)]$ with minimum variance.
\end{definition}

\subsection{Proof of \cref{th:gen_auc_cons}}\label{sec:proof_gen_auc_cons}
Usual arguments imply that:
$L_\lambda(s_\lambda^*) - L_\lambda(\widehat{s}_\lambda) 
\le 2 \cdot \sup_{s\in\mathcal{S}} \abs{ \widehat{L}_\lambda(s) - L_\lambda(s) }$.
Introduce the quantities:
\begin{align*} 
    \widehat{\Delta} = \sup_{s\in\mathcal{S}} 
    \abs{ \widehat{\auc}_{H_s,G_s} - \auc_{H_s,G_s} }
    , \qquad 
    & \widehat{\Delta}_0 = \sup_{s\in\mathcal{S}} 
    \abs{ 
	\widehat{\auc}_{ H_s^{(0)}, G_s^{(0)}} - \auc_{ H_s^{(0)}, G_s^{(0)} }
    } , \\
    \text{and} \quad & 
    \widehat{\Delta}_1 = \sup_{s\in\mathcal{S}} 
    \abs{ 
	\widehat{\auc}_{ H_s^{(1)}, G_s^{(1)} } - \auc_{ H_s^{(1)}, G_s^{(1)} }
    }.
\end{align*} 
The triangular inequality implies that:
$\sup_{s\in\mathcal{S}} \abs{ \widehat{L}_\lambda(s) - L_\lambda(s) }
\le \widehat{\Delta} + \lambda \widehat{\Delta}_0 + \lambda \widehat{\Delta}_1$.

\emph{Case of $\widehat{\Delta}$:} Note that:
\begin{align*}
    \widehat{\auc}_{H_s,G_s} &= (n(n-1)/2n_+n_-)\cdot \widehat{U}_K(s), \\ \text{where} \quad 
    \widehat{U}_K(s) & = \frac{2}{n(n-1)} 
    \sum_{1 \le i < j \le n} K( (s(X_i), Y_i, Z_i), (s(X_j), Y_j, Z_j)),
\end{align*}
and $K( (t,y,z) , (t', y', z')) = \I\{ (y-y')(t-t') > 0\} + (1/2) \cdot \I\{ y \ne y', t=t'\}$.
The quantity $\widehat{U}_K(s)$ is a known type of statistic and is called a $U$-statistic,
see \cref{def:U-statistic} for the definition and \cite{Lee1990} for an overview.
We write $U_K(s):=\E[\widehat{U}_K(s)] = 2 p(1-p)\auc_{H_s,G_s}$.

Following \cite{Clemencon08Ranking}, we have the following lemma.

\begin{lemma}\label{lem:Ubounds} 
    (\citealp{Clemencon08Ranking}, Corollary~3) Assume that $\mathcal{S}$
    is a VC-major class of functions (see \cref{def:VC-major}) with finite {\sc VC} dimension
    $V<+\infty$. We have w.p. $ \ge 1-\delta$: $\forall n>1$,
    \begin{equation}\label{eq:Ubounds}
    \sup_ {S\in \mathcal{S}}
    \left\vert\widehat{U}_K(s) - U_K(s) \right\vert 
    \leq 2C\sqrt{\frac{V}{n}}+2 \sqrt{\frac{\log (1/\delta)}{n-1}},
    \end{equation}
    where $C$ is a universal constant, explicited in \cite{Bousquet2004} (page
    198 therein).
\end{lemma}

Introducing $\widehat{m} := n_+ n_-/n^2 - p(1-p)$, we have that,
since $\sup_{s\in\mathcal{S}} | \widehat{U}_K(s) | \le 2 n_+ n_-/(n(n-1))$:
\begin{align*}
    \widehat{\Delta} 
    & \le \abs{ \frac{n(n-1)}{2n_+n_-} - \frac{1}{2 p(1-p)} } \cdot
    \sup_{s\in\mathcal{S}} \abs{ \widehat{U}_K(s) }
    + \frac{1}{2p(1-p)} \cdot \sup_{s\in\mathcal{S}} \abs{ \widehat{U}_K(s) - U_K(s) }, \\
    & \le \frac{1}{p(1-p)} \abs{ \widehat{m} + \frac{n_+n_-}{n^2(n-1)}} 
    + \frac{1}{2p(1-p)} \cdot \sup_{s\in\mathcal{S}} \abs{ \widehat{U}_K(s) - U_K(s) }.
\end{align*}

The properties of the shatter coefficient described in
\cite{Gyorfi2002} (Theorem 1.12 therein) and the fact that $\mathcal{S}$ is VC major,
imply that the class of sets:
$ \{ ((x,y), (x',y')) \mid (s(x) - s(x'))(y-y') > 0
\}_{s \in \mathcal{S}}$
is VC with dimension $V$.

The right-hand side term above is covered by \cref{lem:Ubounds}, and we deal now with
the left-hand side term.

Hoeffding's inequality implies, that w.p. $\ge 1-\delta$, we have that, for all $n\ge 1$,
\begin{align}\label{eq:Hoeffding_1}
    \abs{ \frac{n_+}{n} - p} \le \sqrt{ \frac{\log(\frac{2}{\delta})}{2n} }.
\end{align}
Since $n_- = n - n_+$, we have that:
\begin{align*}
    \widehat{m} = (1-2 p) \left( \frac{n_+}{n} - p  \right)
    - \left( \frac{n_+}{n} - p \right)^2.
\end{align*}
It follows that:
\begin{align*}
    \abs{ \widehat{m} + \frac{n_+n_-}{n^2(n-1)}} 
    \le \abs{\widehat{m}} + \frac{1}{4(n-1)} 
    \le \left( 1-2p \right) \sqrt{ \frac{\log(2/\delta)}{2n} } + A_n(\delta),
\end{align*}
where $A_n(\delta) = \frac{\log(2/\delta)}{2n} + \frac{1}{4(n-1)} = O\left( n^{-1} \right)$. 

Finally, a union bound between \cref{eq:Ubounds} and \cref{eq:Hoeffding_1} gives that, 
using the 
majoration $1/(2n) \le 1/(n-1)$:
w.p. $\ge 1-\delta$, for any $n > 1$:
\begin{align}\label{eq:delta-result}
    p(1-p) \cdot \widehat{\Delta} \le 
    C \sqrt{\frac{V}{n}}
    + 2(1-p) \sqrt{\frac{\log(3/\delta)}{n-1}} 
    + A_n(2\delta/3).
\end{align}

\emph{Case of $\widehat{\Delta}_0$:} Note that:
\begin{align*}
    \widehat{\auc}_{ H_s^{(0)}, G_s^{(0)} }
    &= \left(n(n-1)/2n_+^{(0)}n_-^{(0)} \right)\cdot \widehat{U}_{K^{(0)}}(s), 
\end{align*}
where $K^{(0)}( (t,y,z) , (t', y', z')) = \I\{ z=0, z'=0\}\cdot K( (t,y,z), (t', y', z') )$.
We denote:
\begin{align*}
    U_{K^{(0)}}(s)
    := \E[\widehat{U}_{K^{(0)}}(s) =  2 q_0^2 p_0 (1-p_0) \cdot \auc_{H_s^{(0)}, G_s^{(0)}}.
\end{align*}
Following the proof of the bound for $\widehat{\Delta}$, introducing
$\widehat{m}_0 := n_+^{(0)}n_-^{(0)} / n^2
- q_0^2 p_0 (1 - p_0)$, 
\begin{align*}
    \widehat{\Delta}_0 
    & \le \frac{1}{q_0^2p_0(1-p_0)} \abs{ \widehat{m}_0 + \frac{n_+^{(0)}n_-^{(0)}}{n^2(n-1)} }
    + \frac{1}{2 q_0^2p_0(1-p_0)} \cdot
    \sup_{s\in \mathcal{S}} \abs{ \widehat{U}_{K^{(0)}}(s) - U_{K^{(0)}}(s) }.
\end{align*}
The right-hand side term above is once again covered by Lemma 1. 
We deal now with the left-hand side term, note that:
\begin{align*}
    \widehat{m}_0 = \; & \frac{n_+^{(0)} n^{(0)}}{n^2} - q_0^2 p_0 
    -  \left( \left[ \frac{n_+^{(0)}}{n} \right]^2 - q_0^2 p_0^2 \right),\\
   = \; & q_0 p_0 \left( \frac{n^{(0)}}{n} - q_0 \right)
    + q_0(1-2p_0) \left( \frac{n_+^{(0)}}{n} - q_0 p_0 \right) \\
    & 
    + \left( \frac{n_+^{(0)}}{n} - q_0 p_0 \right) \left( \frac{n^{(0)}}{n} - q_0 \right)
    - \left( \frac{n_+^{(0)}}{n} - q_0 p_0 \right)^2.
\end{align*}
A union bound of two Hoeffding inequalities gives that
for any $n > 1$, w.p. $\ge 1-\delta$ we have simultaneously:
\begin{align}
    \abs{\frac{n^{(0)}}{n} - q_0} \le \sqrt{\frac{\log(4/\delta)}{2n}} 
    \quad \text{and} \quad 
    \abs{\frac{n_+^{(0)}}{n} - q_0 p_0} \le \sqrt{\frac{\log(4/\delta)}{2n}}.
    \label{proofeq:hoeffding2}
\end{align}
It follows that:
\begin{align*}
    \abs{ \widehat{m}_0 + \frac{n_+^{(0)}n_-^{(0)}}{n^2(n-1)} }
    &\le \abs{\widehat{m}_0} + \abs{\frac{\left(n^{(0)}\right)^2}{4 n^2(n-1)}}
    \le q_0(1-p_0) \sqrt{\frac{\log(4/\delta)}{2n}} + B_n(\delta),
\end{align*}
where $B_n(\delta) = \frac{1}{4(n-1)} + \frac{\log(4/\delta)}{n}.$

Finally, a union bound between \cref{eq:Ubounds} and \cref{proofeq:hoeffding2}
gives,
using the 
majoration $1/(2n) \le 1/(n-1)$, for any $n > 1$:
w.p. $\ge 1-\delta$,
\begin{align}\label{eq:conc_delta_0}
    q_0^2 p_0 (1-p_0) \cdot \widehat{\Delta}_0 
    & \le C\sqrt{\frac{V}{n}} + \left( 1 + q_0(1-p_0) \right)\sqrt{\frac{\log(5/\delta)}{n}}
    + B_n(4\delta/5).
\end{align}

\emph{Case of $\widehat{\Delta}_1$:} 

One can prove a similar result as \cref{eq:conc_delta_0} for $\widehat{\Delta}_1$:
for any $n > 1$: w.p. $\ge 1-\delta$,
\begin{align}\label{eq:conc_delta_1}
    q_1^2 p_1 (1-p_1) \cdot \widehat{\Delta}_1 
    & \le C\sqrt{\frac{V}{n}} + \left( 1 + q_1(1-p_1) \right)\sqrt{\frac{\log(5/\delta)}{n}}
    + B_n(4\delta/5).
\end{align}

\emph{Conclusion:} 

Under the assumption $\min_{ z \in \{0,1\}} \min_{y\in\{-1,1\}} \p\{ Y= y , Z=z \} \ge \epsilon$,
note that $\min(p, 1-p) \ge 2\epsilon$.
A union bound between \cref{eq:delta-result},
\cref{eq:conc_delta_0}, and \cref{eq:conc_delta_1}.
gives that, for any $\delta > 0$ and for all $n > 1$: w.p. $\ge 1 - \delta$,
\begin{align*}
      \epsilon^2 \cdot \left( L_\lambda(s_\lambda^*) - L_\lambda(\widehat{s}_\lambda) \right)
     & \le \;  C\sqrt{\frac{V}{n}} \cdot
     \left( 4\lambda + \frac{1}{2} \right)
     + \sqrt{\frac{\log(13/\delta)}{n-1}} \cdot
     \Big( 4\lambda  + (4\lambda + 2) \epsilon \Big )
     + O(n^{-1}),
\end{align*}
which concludes the proof.


\subsection{Proof of \cref{th:gen_fair_roc}}

Usual arguments imply that:
$L_\Lambda(s_\Lambda^*) - L_\Lambda(\widehat{s}_\Lambda) \le 
2 \cdot \sup_{s\in\mathcal{S}} \abs{ \widehat{L}_\Lambda(s) - L_\Lambda(s) }$.
As in \cref{sec:proof_gen_auc_cons}, the triangle inequality implies that:
\begin{align*}
    & \abs{ \widehat{L}_\Lambda(s) - L_\Lambda(s) } \\
    & \le \abs{\widehat{\auc}_{H_s,G_s} - \auc_{H_s,G_s}} 
    +  \sum_{k=1}^{m_H} \lambda_H^{(k)}
    \abs{ \abs{\widehat{\Delta}_{H,\alpha_k}(s)} - \abs{\Delta_{H,\alpha_k}(s)} }
    + \sum_{k=1}^{m_G} \lambda_G^{(k)} \abs{ 
    \abs{\widehat{\Delta}_{G,\alpha_k}(s) } - \abs{\Delta_{G,\alpha_k}(s)} }, \\
    & \le \abs{\widehat{\auc}_{H_s,G_s} - \auc_{H_s,G_s}} 
    + \sum_{k=1}^{m_H} \lambda_H^{(k)}
    \abs{\widehat{\Delta}_{H,\alpha_k}(s) -\Delta_{H,\alpha_k}(s) }
    + \sum_{k=1}^{m_G} \lambda_G^{(k)} \abs{ 
    \widehat{\Delta}_{G,\alpha_k}(s) - \Delta_{G,\alpha_k}(s)}. 
\end{align*}
It follows that:
\begin{align*}
    \sup_{s\in\mathcal{S}} \abs{ \widehat{L}_\Lambda(s) - L_\Lambda(s) } 
    & \le \sup_{s\in\mathcal{S}} \; \abs{\widehat{\auc}_{H_s,G_s} - \auc_{H_s,G_s}} 
    + \bar{\lambda}_H \cdot \sup_{s, \alpha \in\mathcal{S}\times [0,1]}
    \abs{ \widehat{\Delta}_{H,\alpha}(s) - \Delta_{H,\alpha}(s) } \\
    & \qquad + \bar{\lambda}_G \cdot \sup_{s, \alpha \in\mathcal{S}\times [0,1]} 
    \abs{ \widehat{\Delta}_{G,\alpha}(s) - \Delta_{G,\alpha}(s) },
\end{align*}
and each of the terms is studied independently.
The first term is already dealt with in \cref{sec:proof_gen_auc_cons},
and the second and third terms have the same nature, hence
we choose to focus on $\widehat{\Delta}_{G,\alpha}(s) - \Delta_{G,\alpha}(s) $.

Note that: 
\begin{align*}
    & \widehat{\Delta}_{G,\alpha}(s) - \Delta_{G,\alpha}(s),\\
    & = \widehat{\roc}_{G_s^{(0)}, G_s^{(1)}}(\alpha) - \roc_{G_s^{(0)}, G_s^{(1)}}(\alpha),\\
    & =  \left[ G_{s}^{(1)} \circ \left( G_{s}^{(0)} \right)^{-1}
    - \widehat{G}_{s}^{(1)} \circ \left( \widehat{G}_s^{(0)} \right)^{-1} \right] (1-\alpha), \\
    & = 
   \underbrace{ \left[ 
        G_{s}^{(1)} \circ \left( G_{s}^{(0)} \right)^{-1} 
	-  G_{s}^{(1)}  \circ \left( \widehat{G}_s^{(0)} \right)^{-1}
   \right](1-\alpha) 
   }_{T_1(s,\alpha)} + 
    \underbrace{ \left[
    G_{s}^{(1)} \circ \left( \widehat{G}_s^{(0)} \right)^{-1}  
    - \widehat{G}_{s}^{(1)} \left( \widehat{G}_s^{(0)} \right)^{-1}
    \right](1-\alpha)
    }_{T_2(s,\alpha)}.
\end{align*}
Hence:
\begin{align*}
    \sup_{s,\alpha\in\mathcal{S}\times[0,1]}
    \abs{ \widehat{\Delta}_{G,\alpha}(s) - \Delta_{G,\alpha}(s) }
    \le \sup_{s, \alpha \in\mathcal{S} \times [0,1]} \abs{T_1(s,\alpha)}
    + \sup_{s, \alpha \in\mathcal{S} \times [0,1]} \abs{T_2(s,\alpha)},
\end{align*}
and we study each of these two terms independently.

\emph{Dealing with $\sup_{s, \alpha \in \mathcal{S} \times [0,1]} \abs{T_1(s,\alpha)}$.}

Introduce the following functions, for any $z \in \{0, 1\}$:
\begin{align*}
    \widehat{U}_{n,s}^{(z)}(t) := \frac{1}{n} 
    \sum_{i=1}^n \I\{ Y_i=+1, Z_i=z, s(X_i) \le t\}
    \quad \text{and} \quad U_{n,s}^{(z)}(t) := \E\left[\widehat{U}_{n, s}^{(z)}(t)\right],
\end{align*}
then $\widehat{G}_s^{(z)}(t) = (n/n_+^{(z)}) \cdot \widehat{U}_{n,s}^{(z)}(t)$
and $G_s^{(z)}(t) = (1/q_z p_z) \cdot U_{n,s}^{(z)}(t)$ for any $t \in (0,T)$.

The properties of the generalized inverse of a composition of functions
(see \cite{Vaart2000}, Lemma 21.1, page 304 therein)
give, for any $u \in [0,1]$:
\begin{align}\label{eq:generalized-inverse-1}
    \left( \widehat{G}_s^{(0)} \right)^{-1} (u)  = \left(\widehat{U}_{n,s}^{(0)} \right)^{-1}
    \left( \frac{n_+^{(0)} u}{n} \right).
\end{align}

The assumption on $\mathcal{K}$
implies that $G_s^{(0)}$ is increasing. Define $k^{(0)}_s = G_s^{(0)} \circ s$,
for any $t \in (0, T)$, we have:
\begin{align}\label{eq:k_s_prop}
    \widehat{U}_{n,s}^{(0)}(t) = \widehat{U}_{n, k^{(0)}_s} \left( G_s^{(0)}(t) \right).
\end{align}

Combining \cref{eq:generalized-inverse-1} and \cref{eq:k_s_prop},
we have, for any $u\in[0,1]$:
\begin{align*}
    \left( \widehat{G}_s^{(0)} \right)^{-1} (u)  = 
    \left( G_s^{(0)} \right)^{-1} \circ \left(\widehat{U}_{n,k^{(0)}_s}^{(0)} \right)^{-1}
    \left( \frac{n_+^{(0)}u}{n} \right).
\end{align*}
Since $G_s^{(0)}$ is continuous and increasing, the inverse function theorem
implies that $(G_s^{(0)})^{-1}$ is differentiable.
It follows that:
\begin{align*}
    \frac{d}{du} \left(  G_s^{(1)} \circ (G_s^{(0)})^{-1}(u) \right)
    = \frac{ \left( G_s^{(1)} \right)'\left( (G_s^{(0)})^{-1}(u) \right)}{
    \left( G_s^{(0)} \right)' \left( (G_s^{(0)})^{-1}(u) \right)}
    \le \frac{B}{b},
\end{align*}
and the mean value inequality implies:
\begin{align*}
    \sup_{s, \alpha \in \mathcal{S} \times [0,1]} \abs{T_1(s,\alpha)} 
    & \le (B/b) \cdot \sup_{s, \alpha \in \mathcal{S} \times [0,1]} 
    \abs{ \left(\widehat{U}_{n,k^{(0)}_s}^{(0)} \right)^{-1}
    \left( \frac{n_+^{(0)} \alpha }{n} \right) - \alpha }.
 \end{align*}
Conditioned upon the $Z_i$'s and $Y_i$'s, the quantity
\begin{align*}
    \sqrt{n} \left( \left( \frac{n}{n_+^{(0)}} \right) 
    \hat{U}_{n,k_s^{(0)}}^{(0)}(\alpha) - \alpha\right),
\end{align*}
is a standard empirical process, and it follows from 
\cite{Shorack1989} (page 86 therein), that:
\begin{align*}
     \sup_{\alpha \in [0,1]}
     \abs{
     \left(\widehat{U}_{n,k^{(0)}_s}^{(0)} \right)^{-1}
     \left(  \frac{n_+^{(0)} \alpha }{n} \right) - \alpha}
     = 
     \sup_{\alpha \in [0,1]}
     \abs{
     \frac{n}{n_+^{(0)} }
 \widehat{U}_{n,k^{(0)}_s}^{(0)} \left( \alpha \right) - \alpha }.
\end{align*}

Similar arguments as those seen in \cref{sec:proof_gen_auc_cons} imply:
\begin{align*}
    \sup_{s, \alpha \in \mathcal{S} \times [0,1]} \abs{T_1(s,\alpha)} 
     & \le (B/b) \cdot 
    \sup_{s, \alpha \in \mathcal{S} \times [0,1]} 
     \abs{ \frac{n}{n_+^{(0)} }
     \widehat{U}_{n,k^{(0)}_s}^{(0)} \left( \alpha \right) - \alpha }, \\
     & \le 
      \frac{B}{bq_0p_0} \cdot \abs{ \frac{n_+^{(0)}}{n} - q_0 p_0 }
     +  \frac{B}{bq_0p_0}
	 \cdot 
    \sup_{s, \alpha \in \mathcal{S} \times [0,1]} \abs{ 
	 \widehat{U}_{n,k^{(0)}_s}^{(0)} \left( \alpha \right) - q_0 p_0 \alpha },
\end{align*}


A standard learning bound (see \cite{Boucheron2005}, Theorem 3.2 and 3.4 page
326-328 therein)
implies that: for any $\delta > 0, n > 0$, w.p. $\ge 1-\delta$,
\begin{align}\label{eq:learning-bound-u1}
    \sup_{s, \alpha \in \mathcal{S} \times [0,1]} 
    \abs{\widehat{U}_{n, k^{(0)}_s }^{(0)}(\alpha) - U_{n, k^{(0)}_s}^{(0)}(\alpha)}
    \le C \sqrt{ \frac{V}{n} } + \sqrt{ \frac{2\log(2/\delta)}{n}},
\end{align}
where $C$ is a universal constant.

A union bound between \cref{eq:learning-bound-u1} and a standard Hoeffding inequality
for $n_+^{(0)}$ gives: for any $\delta > 0, n > 1$, w.p. $\ge 1 - \delta$,
\begin{align}\label{eq:T2-conclusion}
    \sup_{s\in\mathcal{S}} \abs{T_1(s,\alpha)} 
    \le \frac{BC}{bq_0p_0} \sqrt{\frac{V}{n}}
    + \frac{3B}{bq_0p_0} 
    \sqrt{ \frac{\log(4/\delta)}{2 n} }.
\end{align}

\emph{Dealing with $\sup_{s, \alpha \in\mathcal{S} \times [0,1]} \abs{T_2(s,\alpha)}$.}

We recall that $\widehat{G}_s^{(z)}(t) = (n/n_+^{(z)}) \cdot \widehat{U}_{n,s}^{(z)}(t)$
and $G_s^{(z)}(t) = (1/q_z p_z) \cdot U_{n,s}^{(z)}(t)$ for any $t \in (0,T)$.

First note that, using the same type of arguments as in \cref{sec:proof_gen_auc_cons}:
\begin{align*}
    \sup_{s, \alpha\in\mathcal{S}\times [0,1]} \abs{T_2(s,\alpha)} 
    & \le \sup_{s,t \in\mathcal{S} \times (0,T)} \abs{\widehat{G}_s^{(1)}(t) - G_s^{(1)}(t) }, \\
    & \le \frac{1}{q_1 p_1} \abs{ \frac{n_+^{(1)}}{n} - q_1 p_1 } 
    + \frac{1}{q_1 p_1} \cdot \sup_{s, t \in\mathcal{S}\times(0,T)} 
    \; \abs{\widehat{U}_{n, s}^{(1)}(t) - U_{n,s}^{(1)}(t)}.
\end{align*}

The same arguments as for \cref{eq:learning-bound-u1} apply, which means that:
for any $\delta > 0, n > 0$, w.p. $\ge 1-\delta$,
\begin{align}\label{eq:inter2-T2}
    \sup_{s, t \in\mathcal{S}\times(0,T)}
    \abs{\widehat{U}_{n, s}^{(1)}(t) - U_{n,s}^{(1)}(t)}
    \le C \sqrt{ \frac{V}{n} } + \sqrt{ \frac{2\log(2/\delta)}{n}},
\end{align}
where $C$ is a universal constant.

A union bound of \cref{eq:inter2-T2} and a standard Hoeffding inequality for $n_+^{(1)}$
finally imply that: for any $\delta > 0, n > 1$, w.p. $1-\delta$,
\begin{align}\label{eq:T1-conclusion}
    \sup_{s\in\mathcal{S}} \abs{T_2(s,\alpha)} 
    \le \frac{C}{q_1p_1} \sqrt{\frac{V}{n}}
    + \frac{3}{q_1p_1} 
    \sqrt{ \frac{\log(4/\delta)}{2 n} }.
\end{align}

\textit{Conclusion.}

Combining \cref{eq:T2-conclusion} and \cref{eq:T1-conclusion}, one obtains that:
for any $\delta > 0, n > 1$, w.p. $\ge 1 - \delta$,
\begin{align}\label{eq:conclusion-delta-G}
    \sup_{s,\alpha\in\mathcal{S}\times[0,1]}
    \abs{ \widehat{\Delta}_{G,\alpha}(s) - \Delta_{G,\alpha}(s) }
    \le C \left( \frac{1}{q_1p_1} + \frac{B}{b q_0 p_0} \right) \sqrt{\frac{V}{n}} 
    + \left( \frac{3}{q_1p_1} + \frac{3B}{bq_0p_0}  \right)
    \sqrt{ \frac{\log(8/\delta)}{2n} }.
\end{align}
and a result with similar form can be shown for $\sup_{s,\alpha\in\mathcal{S}\times[0,1]}
\abs{ \widehat{\Delta}_{H,\alpha}(s) - \Delta_{H,\alpha}(s) }$
by following the same steps.

Under the assumption $\min_{ z \in \{0,1\}} \min_{y\in\{-1,1\}} \p\{ Y= y , Z=z \} \ge \epsilon$,
a union bound between \cref{eq:conclusion-delta-G}, its equivalent for $\widehat{\Delta}_{H,\alpha}$ and
\cref{eq:delta-result} gives, with the majoration $1/(2n) \le 1/(n-1)$:
for any $\delta > 0, n > 1$, w.p. $\ge 1 - \delta$, 
\begin{align*}
\epsilon^2 \cdot \left( L_\Lambda(s_\Lambda^*) - L_\Lambda(\widehat{s}_\Lambda)  \right)
   & \le  2\epsilon\left( 1 + 3(\bar{\lambda}_H + \bar{\lambda}_G)
	\left[ 1 + \frac{B}{b} \right] \right)
	\sqrt{ \frac{\log(19/\delta)}{n-1} }  \\
	& \qquad \qquad +  C \left( \frac{1}{2 } + 2 \epsilon (\bar{\lambda}_H + \bar{\lambda}_G) 
	\left[ 1 + \frac{B}{b} \right] \right) 
	\sqrt{ \frac{V}{n} } + O(n^{-1}),
\end{align*}
which concludes the proof.



\section{Additional Experimental Results and Details}

\subsection{Details on the Training Algorithms}\label{subsec:precisions_optim}

\textbf{General principles.}
Maximizing directly $\widehat{L}_\lambda$ by gradient ascent (GA) is
not feasible, since the criterion is not continuous, hence not differentiable. 
Hence, we decided to approximate any indicator function $x \mapsto \I\{ x > 0 \}$
by a logistic function $\sigma: x \mapsto 1/(1+e^{-x})$.

We learn with stochastic gradient descent using batches $\mathcal{B}_N$ of $N$ elements
sampled with replacement in the training set 
$\mathcal{D}_n = \{(X_i, Y_i, Z_i)\}_{i=1}^n$,
with $\mathcal{B}_N = \{ (x_i, y_i, z_i) \}_{i=1}^N$.
We assume the existence of a small validation dataset $\mathcal{V}_m$,
with $\mathcal{V}_m = \{ (x_i^{(v)}, y_i^{(v)}, z_i^{(v)}) \}_{i=1}^m$.
In practice, one splits a total number of instances $n+m$
between the train and validation dataset.

The approximation of $\widehat{\auc}_{H_s,G_s}$ on the batch writes:
\begin{align*}
    \widetilde{\auc}_{H_s,G_s} = \frac{1}{N_+ N_-} \sum_{i<j} 
    \sigma\left[ (s(x_i)-s(x_j))(y_i-y_j) \right] ,
\end{align*}
where $N_+ := \sum_{i=1}^N \I\{ y_i = +1 \} =: N - N_-$ is the number of
positive instances in the batch.
Similarly, we denote by
$N_+^{(z)} := 
N^{(z)} - N_-^{(z)}$ the number of positive instances of group $z$ in the
batch, with
\begin{align*}
    N^{(z)} := \sum_{i=1}^N \I\{ z_i=z\}
    \quad \text{and} \quad 
    N_+^{(z)} :=
    \sum_{i=1}^N \I\{ z_i = z, y_i = +1 \}.
\end{align*}
Due to the high number of term involved involved in the summation,
the computation of $\widetilde{\auc}_{H_s,G_s} $ can be very expensive,
and we rely on approximations called \emph{incomplete U-statistics}, 
which simply average a random sample of $B$ nonzero terms
of the summation, see \cite{Lee1990}.
We refer
to \cite{JMLRincompleteUstats,Papa2015a} for details on their statistical
efficiency and use in the context of SGD algorithms.
Formally, we define the incomplete approximation with $B\in\N$ pairs
of $\widetilde{\auc}_{H_s,G_s}$ as:
\begin{align*}
    \widetilde{\auc}_{H_s,G_s}^{(B)} := \frac{1}{B} \sum_{(i,j) \in 
    \mathcal{D}_B} 
    \sigma\left[ (s(x_i)-s(x_j))(y_i-y_j) \right] ,
\end{align*}
where $\mathcal{D}_B$ is a random set of $B$ pairs in the set of all possible pairs
$\{(i,j) \mid 1 \le i < j \le N \}$.

\textbf{For $\auc$-based constraints (\cref{subsec:learning-under-auc-cons}).}
Here, we give more details on our algorithm for the case of the $\auc$-based
constraint \cref{rk-cons:intra-pairwise}.
The generalization to other $\auc$-based fairness constraints is straightforward.
For any $z\in\{0,1\}$ the relaxation
of $\widehat{\auc}_{ H^{(z)}, G^{(z)} }$ on the batch writes:
\begin{align*}
    & \widetilde{\auc}_{ H^{(z)}_s, G^{(z)}_s } = 
     \frac{1}{N_+^{(z)} N_-^{(z)}} 
    \sum_{\substack{i < j\\ z_i = z_j = z}}
    \sigma\left[ (s(x_i)-s(x_j))(y_i-y_j) \right].
\end{align*}
Similarly as $\widetilde{\auc}_{H_s, G_s}$, we introduce the sampling-based 
approximations $\widetilde{\auc}_{H^{(z)}_s, G^{(z)}_s}^{(B)}$ for any $z \in \{ 0,1 \}$.

To minimize the absolute value in \cref{eq:auc_general_problem}, we introduce a
parameter $c\in[-1, +1]$, which is modified slightly
every $n_\text{adapt}$ iterations so that it has the same sign as the
evaluation of $\Gamma^\top C(s)$ on $\mathcal{V}_m$.
This allows us to write a cost in the form of a weighted sum of approximated $\auc$'s, 
with weights that vary during the optimization process.
Precisely, it is defined as:
\begin{align*}
    \widetilde{L}_{\lambda, c}(s) :=
    \left(1-  \widetilde{\auc}_{H_s,G_s}\right)
    + \lambda \cdot c \left( \widetilde{\auc}_{H^{(1)}_s, G^{(1)}_s}
    - \widetilde{\auc}_{H^{(0)}_s, G^{(0)}_s}
\right) + 
\frac{\lambda_\text{reg}}{2}
\cdot \norm{W}_2^2,
\end{align*}
where $\lambda_\text{reg}$ is a regularization parameter and $\norm{W}_2^2$
is the sum of the squared $L_2$ norms of all of the weights of the model.
The sampling-based approximation of $\widetilde{L}_{\lambda, c}$ writes:
\begin{align*}
    \widetilde{L}_{\lambda, c}^{(B)}(s) := 
    \left(1- \widetilde{\auc}_{H_s, G_s}^{(B)}\right)
    + \lambda \cdot c \left( \widetilde{\auc}_{H^{(1)}_s, G^{(1)}_s}^{(B)}
    - \widetilde{\auc}_{H^{(0)}_s, G^{(0)}_s}^{(B)}
 \right) + 
\frac{\lambda_\text{reg}}{2}
\cdot \norm{W}_2^2.
\end{align*}
The algorithm is detailed in \cref{alg:sec3}, where $\text{sng}$ is the sign
function,
\textit{i.e.} $\text{sgn}(x) = 2\I\{ x > 0\} -1$ for any $x \in \R$.

\begin{algorithm}[h]
    \setstretch{1.20}
    \caption{Practical algorithm for learning with the $\auc$-based constraint \cref{rk-cons:intra-pairwise}.}
   \label{alg:sec3}
\begin{algorithmic}
    \STATE {\bfseries Input:} training set $\mathcal{D}_n$, validation set $\mathcal{V}_m$
    \STATE $c \leftarrow 0$
   \FOR{$i=1$ {\bfseries to} $n_{\text{iter}}$}
   \STATE $\mathcal{B}_N\leftarrow$ $N$ observations sampled with replacement
   from $\mathcal{D}_n$
   \STATE $s\leftarrow$ updated scoring function using a gradient-based
   algorithm (\textit{e.g.} ADAM), using
   the derivative of $\widetilde{L}_{\lambda, c}^{(B)}(s)$ on $\mathcal{B}_N$
   \IF{$(n_\text{iter} \; \text{mod} \; n_{\text{adapt}}  )= 0$}
   \STATE $\Delta\auc \leftarrow 
   \widehat{\auc}_{H_s^{(1)}, G_s^{(1)}}^{(B_{\text{v}})} 
   - \widehat{\auc}_{H_s^{(0)}, G_s^{(0)}}^{(B_{\text{v}})}$
   computed on $\mathcal{V}_m$
   \STATE $c \leftarrow c + \text{sgn}(\Delta\auc) \cdot \Delta c$
   \STATE $c \leftarrow \min(1, \max(-1, c))$
   \ENDIF
   \ENDFOR
   \STATE {\bfseries Output:} scoring function $s$
\end{algorithmic}
\end{algorithm}

\textbf{For $\roc$-based constraints (\cref{sec:sec4-statistical-guarantees}).}
We define an approximation of the quantities $\widehat{H}_s^{(z)},
\widehat{G}_s^{(z)}$ on $\mathcal{B}_N$, for any $z \in \{0, 1\}$, as:
\begin{align*}
    \widetilde{H}_s^{(z)}(t) = \frac{1}{N_-^{(z)}} 
    \sum_{i=1}^N \I\{ y_i = -1, z_i = z\} \cdot 
    \sigma( t - s(x_i)),\\
    \widetilde{G}_s^{(z)}(t) = \frac{1}{N_+^{(z)}} 
    \sum_{i=1}^N \I\{ y_i = +1, z_i = z\} \cdot 
    \sigma( t - s(x_i)).
\end{align*}
which can be respectively seen as relaxations of the false positive rate
(\textit{i.e.} $\widebar{H}_s^{(z)}(t) = 1 - H_s^{(z)}(t)$)
and true positive rate (\textit{i.e.} $\widebar{G}_s^{(z)}(t) = 1 - G_s^{(z)}(t)$)
at threshold $t$ and conditioned upon $Z=z$. 

For any $F \in \{ H,G \}, k \in \{ 1,\dots, m_F \}$,
we introduce a loss $\ell_{F}^k$ which gradients
are meant to enforce the constraint $|\widehat{\Delta}_{ F, \alpha_F^{(k)} }(s)| = 0$.
This constraint can be seen as one that imposes equality between the true positive rates
and false positive rates for the problem of discriminating between the negatives
(resp. positives) of sensitive group 1 against those of sensitive group 0 
when $F=H$ (resp. $F=G$).
An approximation of this problem's false positive rate (resp. true positive rate)
at threshold $t$ is $\widetilde{F}_s^{(0)}(t)$ (resp. $\widetilde{F}_s^{(1)}(t)$).
Introduce $c_F^{(k)}$ as a constant in $[-1, +1]$ and $t_F^{(k)}$
as a threshold in $\R$, the following
loss $\ell_F^{(k)}$ seeks to equalize these two quantities at threshold $t_F^{(k)}$:
\begin{align*}
    \ell_F^{(k)}(s) = 
    c_{F}^{(k)} \cdot \left( \widetilde{F}_s^{(0)} \left( t_F^{(k)} \right)
    - \widetilde{F}_s^{(1)} \left( t_F^{(k)} \right) \right).
\end{align*}
If the gap between 
$\widehat{F}_s^{(0)}(t_F^{(k)})$ and $\widehat{F}_s^{(1)}(t_F^{(k)})$
--- evaluated on the validation set $\mathcal{V}_m$ --- is not too large,
the threshold $t_F^{(k)}$ is modified slightly every few iterations
so that $\widehat{F}_s^{(0)}(t_F^{(k)})$ and
$\widehat{F}_s^{(1)}(t_F^{(k)})$ both approach the target value $\alpha_F^{(k)}$.
Otherwise, the parameter $c_F^{(k)}$ is slightly modified.
The precise strategy to modify $c_F^{(k)}$ and $t_F^{(k)}$ is detailed in \cref{alg:sec4},
and we introduce a step $\Delta t$ to modify the thresholds $t_F^{(k)}$.

The final loss writes:
\begin{align*}
    \widetilde{L}_{\Lambda, c, t}(s) :=
    \left( 1 - \widetilde{\auc}_{H_s,G_s} \right)
    + \frac{1}{m_H} \sum_{k=1}^{m_H} \lambda_H^{(k)} \cdot \ell_H^{(k)} (s)
    + \frac{1}{m_G} \sum_{k=1}^{m_G} \lambda_G^{(k)} \cdot \ell_G^{(k)} (s)
    + 
    \frac{\lambda_\text{reg}}{2} \cdot \norm{W}_2^2, 
\end{align*}
and one can define $\widetilde{L}_{\Lambda, c, t}^{(B)}$
by approximating $\widetilde{\auc}_{H_s,G_s}$ above by $\widetilde{\auc}_{H_s,G_s}^{(B)}$.
The full algorithm is given in \cref{alg:sec4}.

\begin{algorithm}[h]
    \setstretch{1.60}
    \caption{Practical algorithm for learning with $\roc$-based constraints.}\label{alg:sec4}
\begin{algorithmic}
    \STATE {\bfseries Input:} training set $\mathcal{D}_n$, validation set $\mathcal{V}_m$
    \STATE $c_{F}^{(k)}\leftarrow 0$ for any $F\in \{H,G\}$, $k \in \{1, \dots, m_F\}$
    \STATE $t_{F}^{(k)}\leftarrow 0$ for any $F\in \{H,G\}$, $k \in \{1, \dots, m_F\}$
   \FOR{$i=1$ {\bfseries to} $n_{\text{iter}}$}
   \STATE $\mathcal{B}_N\leftarrow$ $N$ observations sampled with replacement
   from $\mathcal{D}_n$
   \STATE $s\leftarrow$ updated scoring function using a gradient-based
   algorithm (\textit{e.g.} ADAM), using 
   the derivative of $\widetilde{L}_{\Lambda, c, t}^{(B)}(s)$ on $\mathcal{B}_N$
   \IF{$(n_\text{iter} \; \text{mod} \; n_{\text{adapt}}  )= 0$}
   \FOR{\textbf{any} $F\in \{H,G\}$, $k \in \{1, \dots, m_F\}$}
   \STATE $\Delta_F^{(k)} \leftarrow
       \widehat{F}_s^{(0)} \left( t_F^{(k)} \right)
	- \widehat{F}_s^{(1)} \left( t_F^{(k)} \right)$ computed on $\mathcal{V}_m$
    \STATE $\Sigma_F^{(k)} \leftarrow
       \widehat{F}_s^{(0)} \left( t_F^{(k)} \right)
       + \widehat{F}_s^{(1)} \left( t_F^{(k)} \right) - 2 \alpha_F^{(k)}$
       computed on $\mathcal{V}_m$
       \IF{$\abs{\Sigma_F^{(k)}} > \abs{\Delta_F^{(k)}} $}
	    \STATE 
	    $ t_F^{(k)} \leftarrow
	    t_F^{(k)} + \text{sgn} \left( \Sigma_F^{(k)} \right) \cdot \Delta t$
	\ELSE
	    \STATE $c_F^{(k)} \leftarrow c_F^{(k)} + \text{sgn}\left(\Delta_F^{(k)} \right)
	    \cdot \Delta c$
	   \STATE $c_F^{(k)} \leftarrow \min\left(1, \max\left(-1, c_F^{(k)}\right)\right)$
	\ENDIF
   \ENDFOR
   \ENDIF
   \ENDFOR
   \STATE {\bfseries Output:} scoring function $s$
\end{algorithmic}
\end{algorithm}

\textbf{Choice of scoring functions and optimization.}
To parameterize the family of scoring functions, we used a simple neural
network of various depth $D$ ($D=0$ corresponds to
a linear scoring function,
while $D=2$ corresponds to a network of 2 hidden layers). Each layer
has the same width $d$ (the dimension of the input space), except for the
output layer which outputs a real score. We used ReLU's as activation functions.
To center and scale the output score
we used \textit{batch normalization} (BN) (see Section 8.7.1 in \cite{Goodfellow2016})
with fixed values $\gamma=1, \beta=0$ for the output value of the network.
\cref{alg:network-architecture} gives a formal description of the network
architecture.
The intuition for normalizing the output score is that the ranking 
losses only depend on the relative value of the score between instances,
and the more \textit{classification-oriented} losses of ROC-based constraints
only depend on a threshold on the score.
Empirically, we observed the necessity of renormalization
for the algorithm with $\roc$-based constraints, as the loss $\ell_F^{(k)}$
is zero when $\widehat{F}_s^{(0)}(t_F^{(k)}) = \widehat{F}_s^{(1)}(t_F^{(k)}) \in \{0, 1\}$,
which leads to scores that drift away from zero during the learning process,
as it seeks to satisfy the constraint imposed by $\ell_F^{(k)}$.
All of the network weigths were initialized using a simple centered normal random
variable with standard deviation $0.01$.

\begin{algorithm}[h]
    \setstretch{1.60}
    \caption{Network architecture.}\label{alg:network-architecture}
\begin{algorithmic}
    \STATE {\bfseries Input:} observation $x = h''_0 \in\R^d$,
   \FOR{$k=1$ {\bfseries to} $D$}
       \STATE \textit{Linear layer:} 
       $h_{k} = W_{k}^\top h''_{k-1} + b_{k}$ with $W_{k} \in \R^{d, d}, 
       b_{k}\in \R^{d, 1}$
       learned by GD,
       \STATE \textit{ReLu layer:} $h''_{k} = \max(0, h'_k)$ where $\max$ is an
       element-wise maximum,
   \ENDFOR
   \STATE \textit{Linear layer:} 
   $h_{D+1} = w_{D+1}^\top h''_{D} + b_{D+1}$ with $w_{D+1} \in \R^{d, 1}, 
   b_{D+1}\in \R$
   learned by GD,
   \STATE \textit{BN layer:} $h'_{D+1} = (h_{D+1} - \mu_{D+1})/\sigma_{D+1}$,
   with $\mu_{D+1}\in \R, \sigma_{D+1} \in \R$ running averages,
   \STATE {\bfseries Output:} score $s(x)$ of $x$, with $s(x) = h'_{D+1} \in\R$.
\end{algorithmic}
\end{algorithm}

For both $\auc$-based and $\roc$-based constraints, optimization was
done with the ADAM algorithm. It features an adaptive step size, so we did
not modify the default parameters. We refer to \cite{Ruder2016} for more
details
on
gradient descent optimization algorithms.

\textbf{Implementation details.}
For all experiments, we set aside $40\%$ of the data for validation,
\textit{i.e.} $m = \lfloor 0.40(m+n) \rfloor$ with $\lfloor \cdot \rfloor$ the floor function,
the batch size to $N=100$ and the parameters 
of the loss changed every $n_\text{adapt} = 50$ iterations.
For any sampling-based approximation computed on a batch $\mathcal{B}_N$,
we set $B = 100$, and $B_{\text{v}}=10^5$ for those on a validation set $\mathcal{V}_m$.
The value $\Delta c$ was always fixed to $0.01$ and $\Delta t$ to $0.001$.
We used linear scoring functions, \textit{i.e.} $D=0$, for the synthetic data
experiments,
and networks with $D=2$ for real data.

The experiments were implemented in Python, and relied extensively
on the libraries \texttt{numpy}, \texttt{TensorFlow} \citep{tensorflow},
\texttt{scikit-learn} \citep{scikit-learn} and
\texttt{matplotlib} for plots.
The code and data can be found in the following repository: 
\url{https://github.com/RobinVogel/Learning-Fair-Scoring-Functions}.

\subsection{Synthetic Data Experiments}

The following examples 
introduce data distributions that we use to illustrate the relevance of our
approach.

\begin{figure}[t]
    \centering
    \includegraphics[width=.45\linewidth]
    {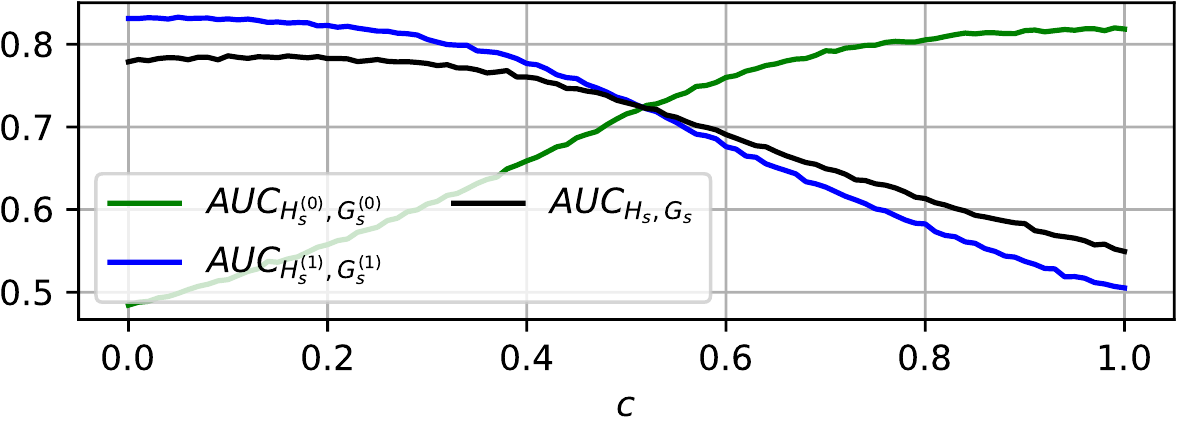}
    \caption{Plotting \cref{example-square} for $q_1 =
    17/20$.
    Under the fairness definition \cref{rk-cons:intra-pairwise}, a fair
    solution exists for $c=1/2$, but the ranking performance for
    $c<1/2$ is significantly higher.}
    \label{fig:example-square}
\end{figure}

\begin{example}\label{example-square}
    Let $\X = [0,1]^2$. For any $x = (x_1 \; x_2)^\top \in \X$, 
    let $\mu^{(0)}(x) = \mu^{(1)}(x) = 1$,
    as well as $\eta^{(0)}(x) = x_1$ and $\eta^{(1)}(x) = x_2$.
    We have $\mu(x) = 1$ and $\eta(x) = q_0 x_1 + q_1 x_2$.
    Consider linear scoring functions of the form $s_c(x) = c
    x_1 + (1-c) x_2$ parameterized by $c \in [0,1]$.
    \cref{fig:example-square} plots $AUC_{H_s, G_s}$
    and $AUC_{H^{(z)}_s, G^{(z)}_s}$ for $z \in \{0,1\}$ as a function of
    $c$, illustrating the trade-off between fairness and ranking performance.
\end{example}

\begin{example}\label{example:toy2}
    Set $\X = [0,1]^2$. For any $x\in\X$ with $x = (x_1 \; x_2)^\top$, 
    set $\mu^{(0)}(x) = (16/\pi) \cdot \I\{ x^2 + y^2 \le 1/2\}$, $\mu^{(1)}
    (x) = (16/3\pi) \cdot \I\{ 1/2 \le x^2 + y^2 \le 1 \}$, and $\eta^{(0)}(x) = \eta^{(1)}(x) = (2/\pi) \cdot \arctan(x_2/x_1)$.
\end{example}

For all of the synthetic data experiments, our objective is to show that the
learning procedure recovers the optimal scoring function when the dataset is
large
enough.
Each of the 100 runs that we perform uses independently generated train,
validation and test datasets.
The variation that we report on 100 runs hence includes that of the data
generation process.
For each run, we chose a total of $n + m=10,000$ points for the train and validation 
sets and a test dataset of size $n_{\text{test}}=20,000$.
Both algorithms ran for $n_\text{iter} = 10,000$ iterations, 
and with the same regularization strength $\lambda_{\text{reg}} = 0.01$.

\textbf{Solving \cref{example-square}.}
First, we illustrate learning with the $\auc$ 
constraint in \cref{rk-cons:intra-pairwise} on the simple problem in 
\cref{example-square}.
Our experiment shows that we can effectively 
find trade-offs between ranking accuracy and satisfying \cref{rk-cons:intra-pairwise}
using the procedure described in \cref{alg:sec3}.

\begin{figure}[h]


    \centering
    \includegraphics[width=0.4\linewidth]{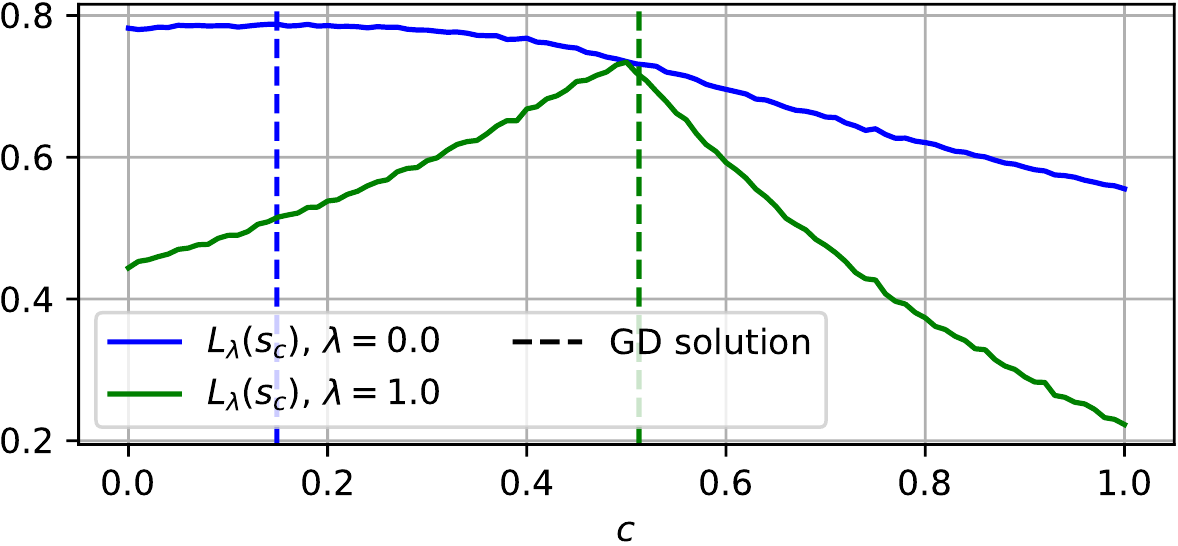}
    \caption{For \cref{example-square}, $L_\lambda(s_c)$ as a function of $c \in [0,1]$ for $\lambda \in \{ 0, 1\}$,
	with the parametrization $s_c(x) = cx_1 + (1-c)x_2$,
	and the values $c$ for the scores obtained by gradient descent with
	\cref{alg:sec3}.
    }\label{fig:solutions-toy1}
    
    \vspace{1.1cm}

%
    \hfill
    \centering
    \raisebox{0.85\height}{\includegraphics[width=0.15\linewidth]{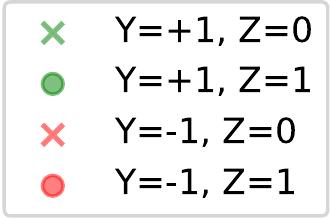}}
    \subfigure[$\lambda=0$]{
    \includegraphics[width=0.33\linewidth]{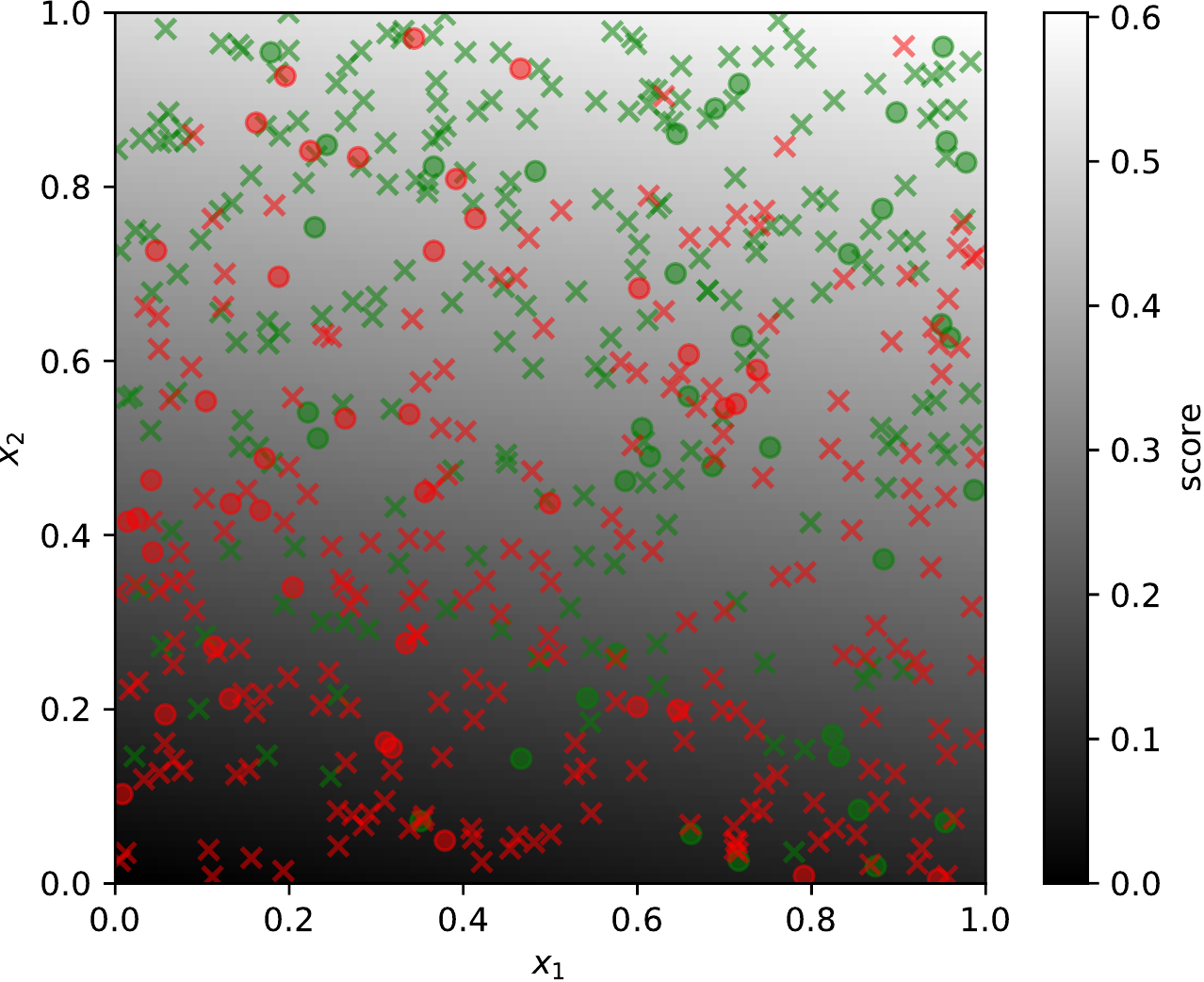}
    }
    \subfigure[$\lambda = 1$]{
    \includegraphics[width=0.33\linewidth]{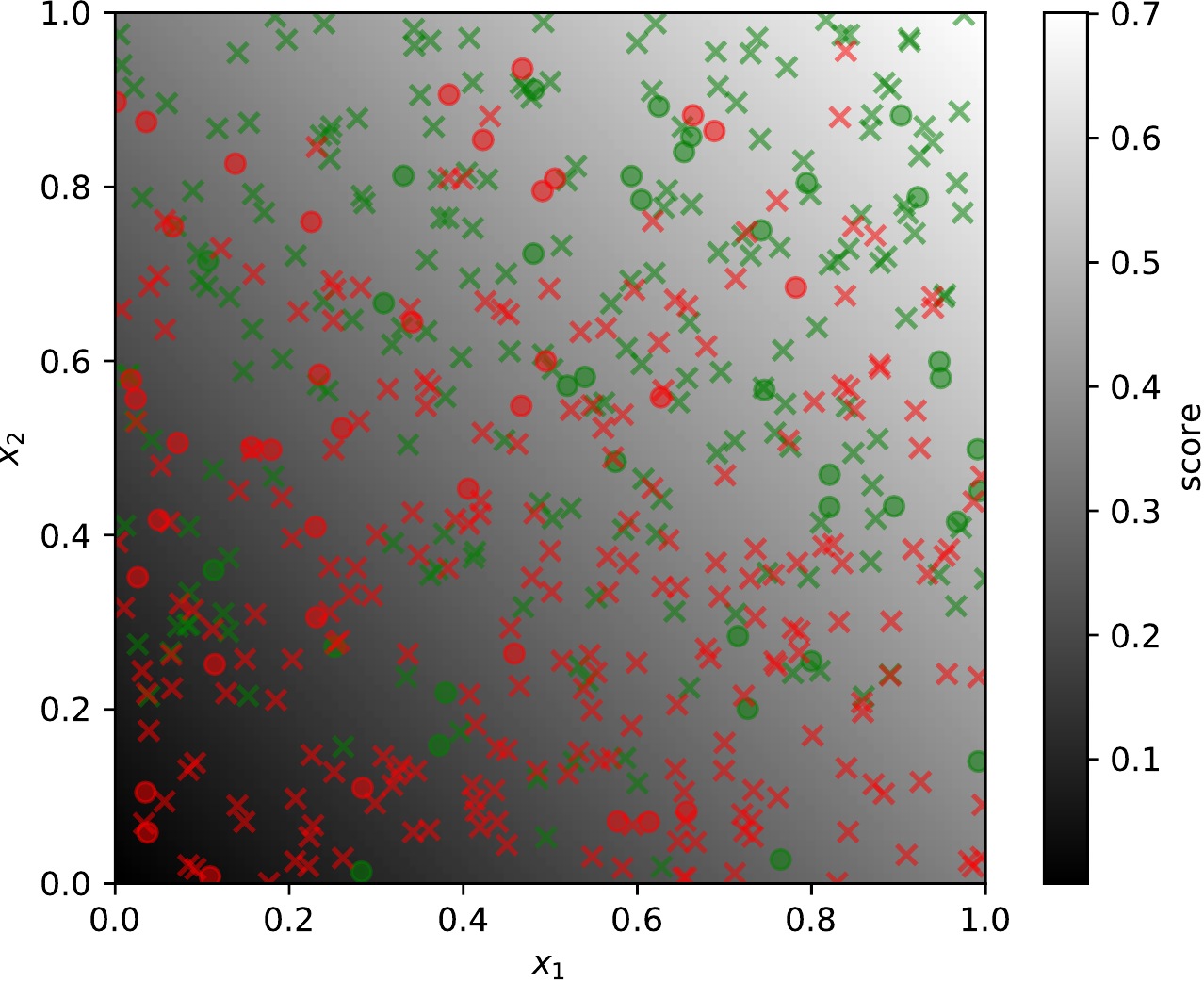}
    }
    \hfill
    \caption{Values of the output scoring functions on $[0,1]^2$
	for \cref{alg:sec3} 
	ran on \cref{example-square}.
    }\label{fig:scores_toy1}
    
    \vspace{1.1cm}

%
\centering
    \includegraphics[width=0.50\columnwidth]{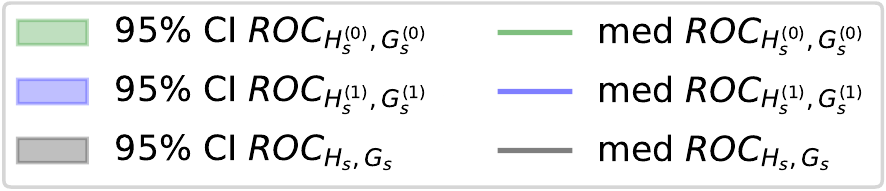} 

    \subfigure[$\lambda = 0$]{\label{fig:ex-square-sol-low-lambda}
    \includegraphics[width=0.25\columnwidth]{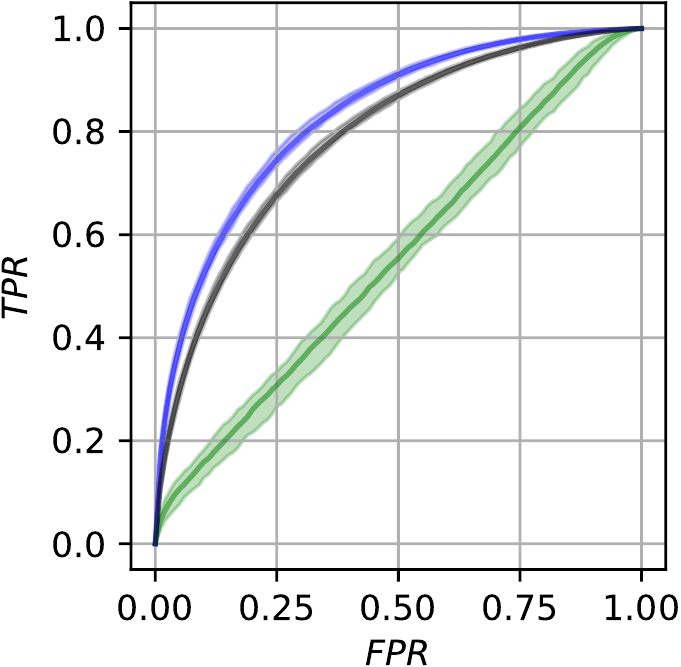}}
    \subfigure[$\lambda = 1$]{\label{fig:ex-square-sol-high-lambda}
    \includegraphics[width=0.25\columnwidth]{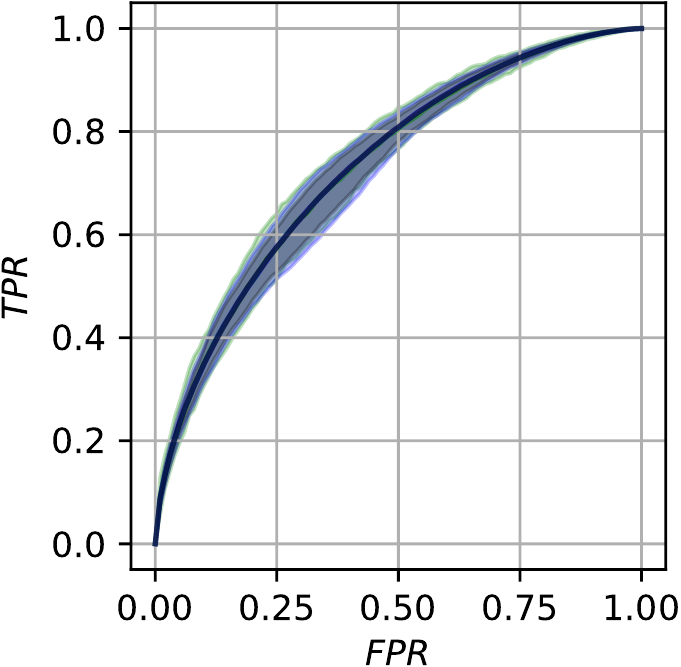}
    }
\caption{Result of \cref{example-square} with \cref{alg:sec3}.}\label{fig:roc-curves-toy1}
\end{figure}

The final solutions of \cref{alg:sec3} with two different values of $\lambda$,
parameterized by $c$, are shown in \cref{fig:solutions-toy1}. 
A representation of the value of the corresponding scoring functions on
$[0,1]\times[0,1]$ is provided in \cref{fig:scores_toy1}.
The median $\roc$ curves for two values of $\lambda$ over 100 independent runs 
are shown in \cref{fig:roc-curves-toy1}, with pointwise 95\% confidence intervals.

\textbf{Solving \cref{example:toy2}.}
\cref{example:toy2} allows to compare 
AUC-based and ROC-based approaches. The former uses 
\cref{rk-cons:intra-pairwise} as constraint and the latter penalizes
$\Delta_{H, 3/4}(s)\ne 0$.
The goal of our experiment with \cref{example:toy2}
is to show that \cref{alg:sec4} can effectively
learn a scoring function $s$ for which the $\alpha$ corresponding to
a classifier $g_{s,t_\alpha}$ that is fair in FPR is specified in advance,
and that the solution can be significantly different from those obtained with
AUC-based constraints and \cref{alg:sec3}.

We compare the solutions of optimizing the $\auc$ without constraint,
\textit{i.e.} \cref{alg:sec3} with $\lambda = 0$ with those of 
\cref{alg:sec3} with $\lambda = 1$ and \cref{alg:sec4} where we impose
$\Delta_{H, 3/4}(s) = 0$ with strength $\lambda_H =1$.
To illustrate the results, we introduce the following family of scoring
functions $s_c(x)
= - c\cdot x_1 + (1-c) \cdot x_2$, parameterized by $c\in[0,1]$.

In practice, we observe that the different constraints lead to scoring
functions with
specific trade-offs between fairness and performance, as summarized in 
\cref{tab:table-all-results-synthetic}.
Results with $\auc$-based fairness are the same for $\lambda = 0$
and $\lambda = 1$ because the optimal scoring function for ranking satisfies
\cref{rk-cons:intra-pairwise}.

\begin{figure}[h]


    \centering
    \includegraphics[width=0.4\linewidth]{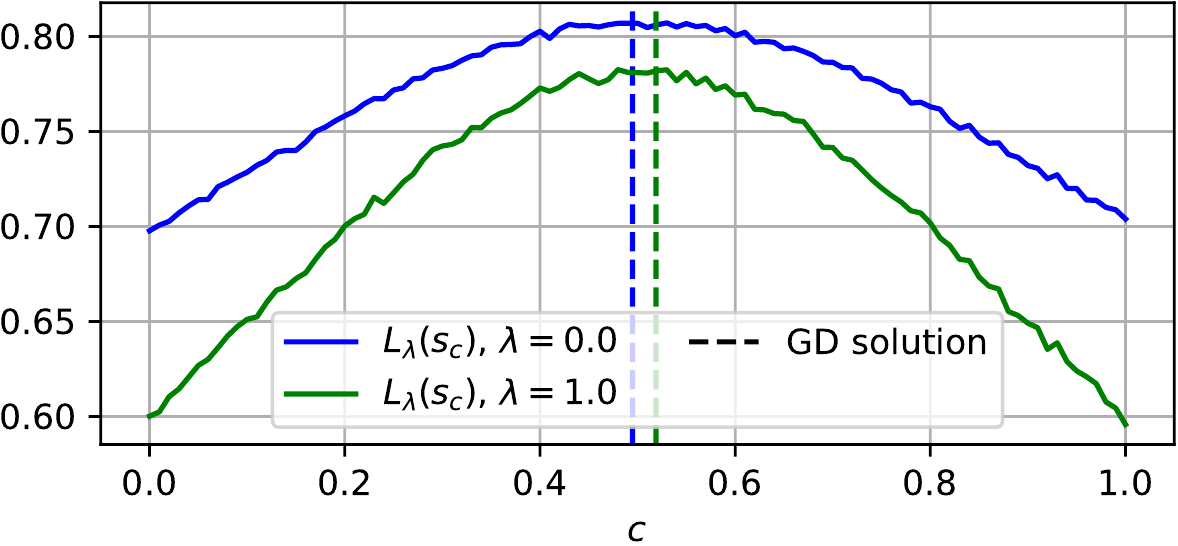}
    \includegraphics[width=0.4\linewidth]{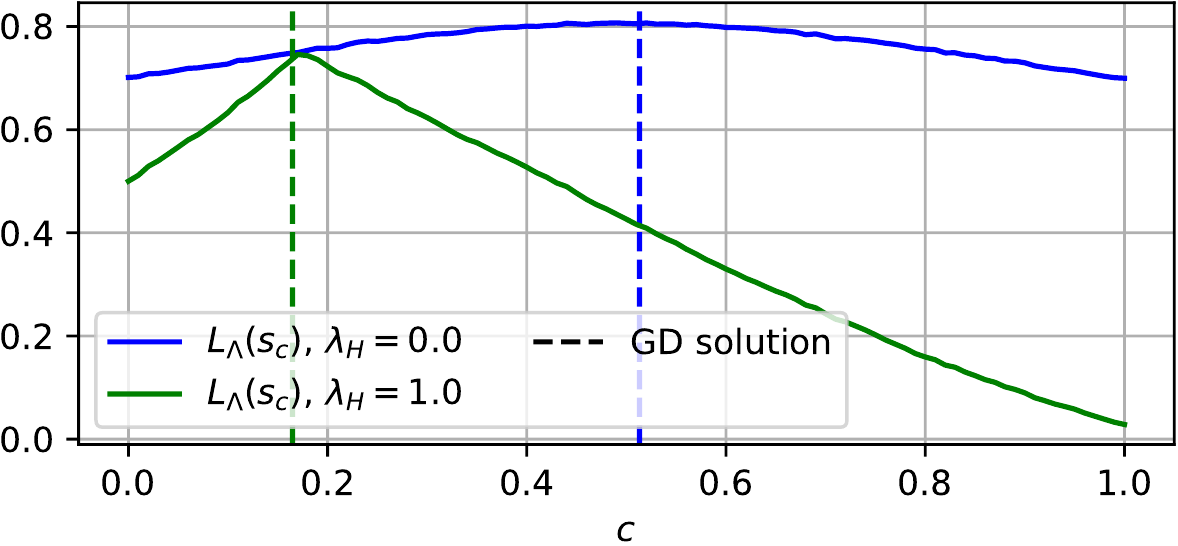}
    \caption{On the left (resp. right), for \cref{example:toy2}, $L_\lambda(s_c)$ (resp. $L_\Lambda(s_c)$) as a function 
	of $c \in [0,1]$ for $\lambda \in \{ 0, 1\}$ (resp. $\lambda_H \in \{0,1\}$),
	with the parametrization $s_c(x) = -cx_1 + (1-c)x_2$,
	and the values $c$ for the scores obtained by gradient descent with
	\cref{alg:sec3} (resp. \cref{alg:sec4}).
    }\label{fig:solutions-toy2}
%
    \vspace{1.1cm}

    \hfill
    \centering
    \raisebox{0.85\height}{\includegraphics[width=0.15\linewidth]{figures/supp/synth_data/score_2d_legend.pdf}}
    \subfigure[$\lambda_H=0$]{
    \includegraphics[width=0.33\linewidth]{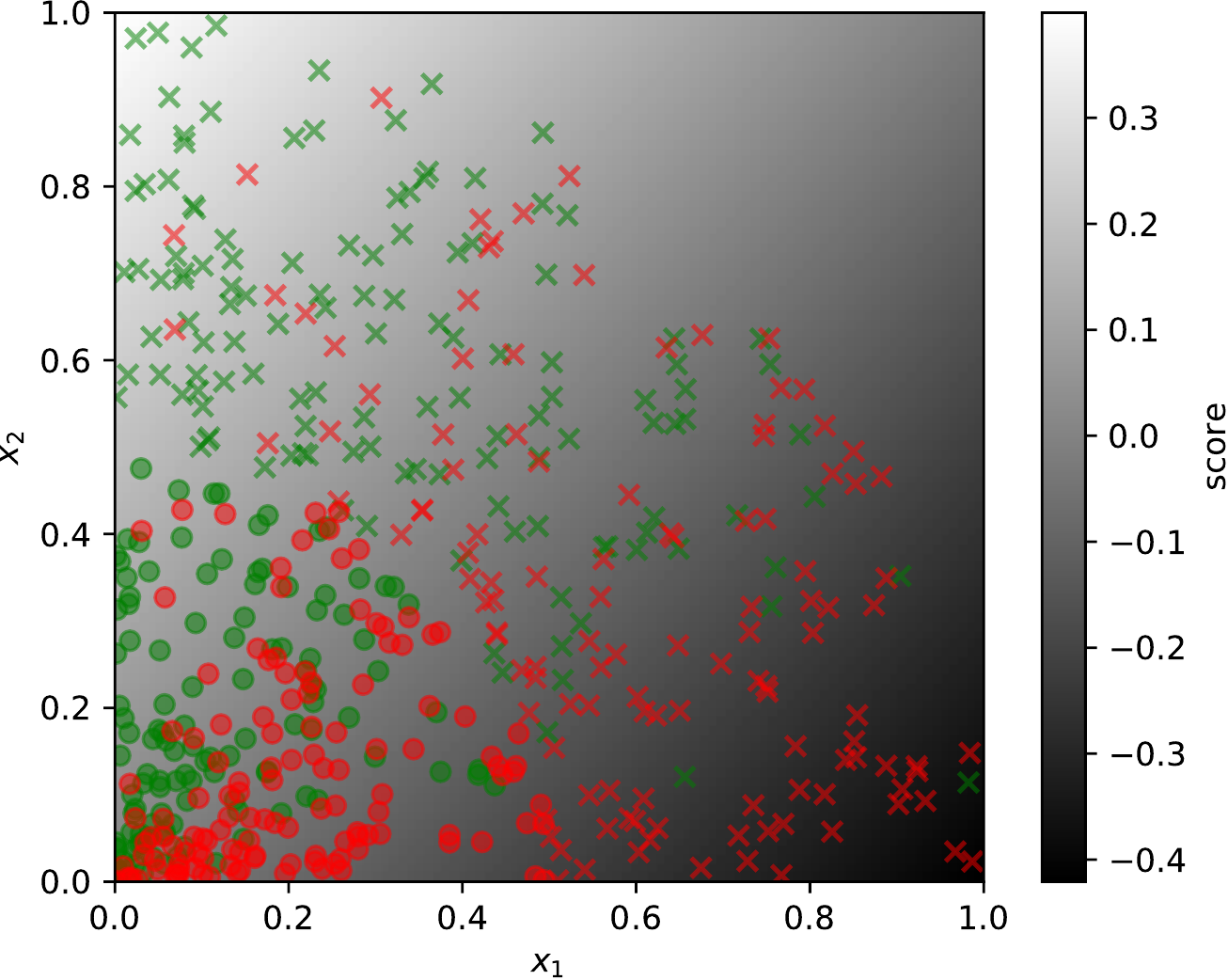}
    }
    \subfigure[$\lambda_H = 1$]{
    \includegraphics[width=0.33\linewidth]{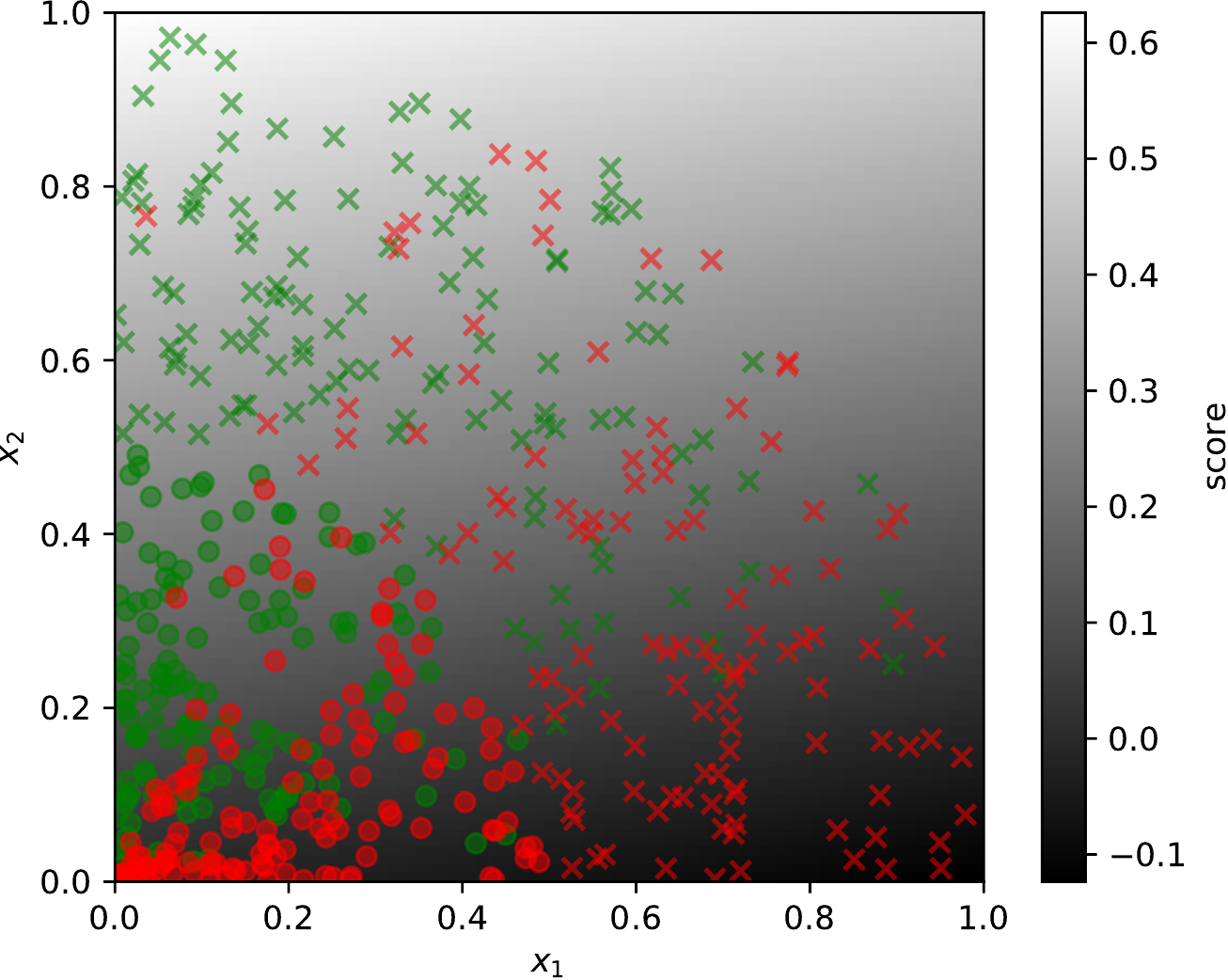}
    }
    \hfill
    \caption{Values of the output scoring functions on $[0,1]^2$
	for \cref{alg:sec4} 
	ran on \cref{example:toy2}.
    }\label{fig:scores_toy2}
%

    \vspace{1.1cm}
\centering
    \includegraphics[width=0.4\columnwidth]{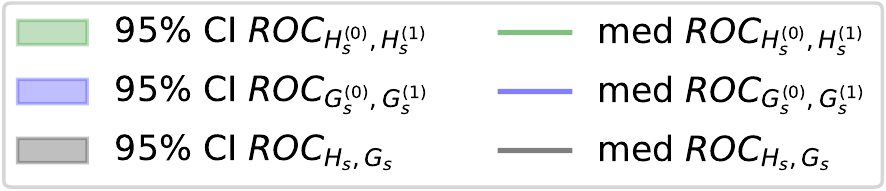} 

    \subfigure[$\lambda_H = 0$]{\label{fig:ex-toy2-sol-low-lambda}
    \includegraphics[width=0.25\columnwidth]{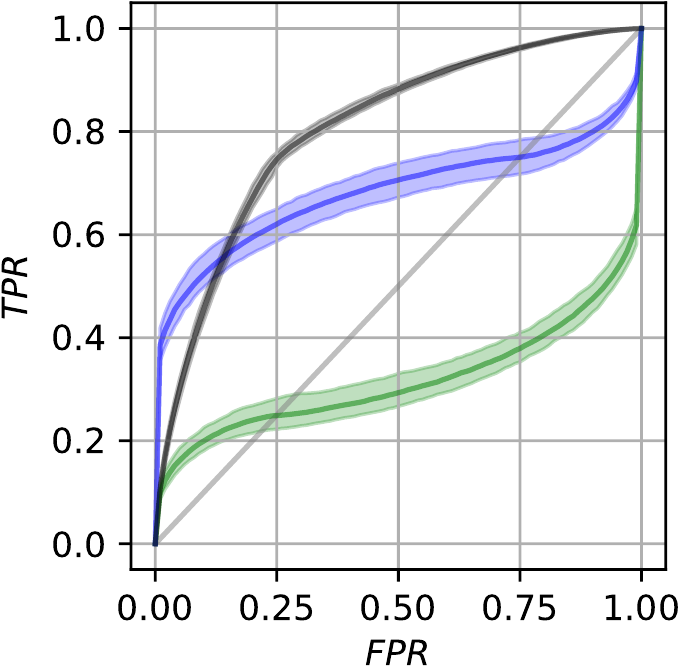}}
    \subfigure[$\lambda_H = 1$]{\label{fig:ex-toy2-sol-high-lambda}
    \includegraphics[width=0.25\columnwidth]{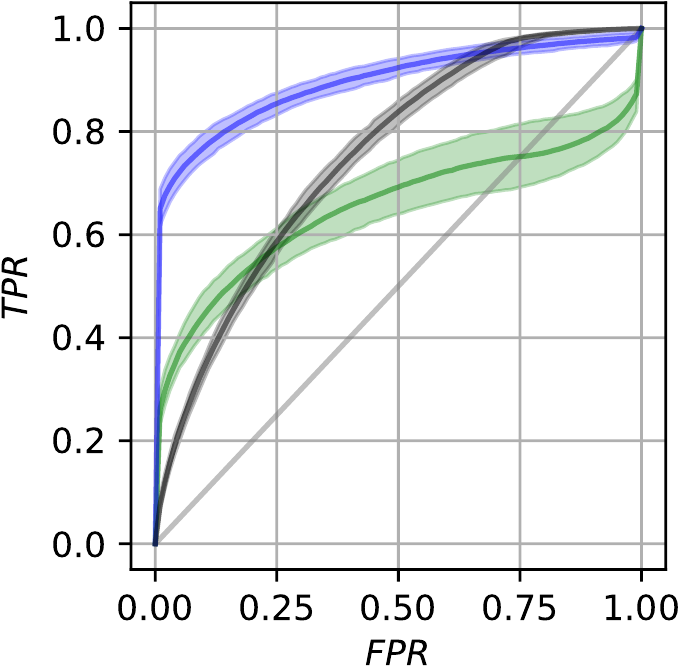}
    }
\caption{Result of \cref{example:toy2} with \cref{alg:sec4}.}\label{fig:roc-curves-toy2}

\end{figure}

\cref{fig:solutions-toy2} shows that the $\auc$-based constraint has no effect on the solution,
unlike the $\roc$-based constraint which is successfully enforced by 
\cref{alg:sec4}. \cref{fig:scores_toy2} gives two possible scoring functions with
\cref{alg:sec4}.
The median $\roc$ curves for two values of $\lambda_H$ over 100
independent runs are shown in \cref{fig:roc-curves-toy1}, with pointwise 95\%
confidence intervals.




\begin{table*}[h]
    \centering
    \caption{Results on the test set, averaged over 100 runs
	(std. dev. are all smaller than $0.02$).
    }\label{tab:table-all-results-synthetic}
    {\footnotesize
    \begin{tabular}{lcccccccc}
  \toprule
  Method & 
  \multicolumn{5}{c}{$\auc$-based fairness} & 
  \multicolumn{3}{c}{$\roc$-based fairness} \\ 
  \cmidrule(lr){1-1}\cmidrule(lr){2-6}\cmidrule(lr){7-9}
  Value of $\lambda$ &  
  \multicolumn{2}{c}{$\lambda=0$} & 
  \multicolumn{3}{c}{$\lambda > 0$} & 
  \multicolumn{3}{c}{$\lambda_H^{(k)} = \lambda_H > 0$} \\
  \cmidrule(lr){1-1}\cmidrule(lr){2-3}\cmidrule(lr){4-6}
  \cmidrule(lr){7-9}
    & 
   $\auc$ & $\Delta \auc$ &
   $\auc$ & $\Delta \auc$ & 
   $|\Delta_{H, 3/4}|$ 
   & $\auc$ & $\Delta \auc$ & 
   $|\Delta_{H, 3/4}|$ \\
   \midrule 
  \cref{example-square} 
  & \textbf{0.79} &  0.28
  & 0.73 & \textbf{0.00} & -- & -- 
  & -- &  -- \\
  \cref{example:toy2} 
  & \textbf{0.80} & \textbf{0.02}
  & \textbf{0.80} & \textbf{0.02} & 0.38 
  & 0.75 & 0.06 & \textbf{0.00} \\ 
  \bottomrule
    \end{tabular}
    }
\end{table*}

\subsection{Real Data Experiments}\label{sec:appendix:real_data_experiments}

\textbf{Datasets.}
We evaluate our algorithms on four datasets that have been commonly used in
the fair machine
learning literature. Those are the following:
\begin{itemize}
    \item The \textit{Compas Dataset} (Compas), featured in
  \citep{Zehlike2017FAIRAF,Donini2018}, consists in predicting recidivism
  of convicts in the US. The sensitive variable is the race of the
  individual, precisely $Z=1$ if the individual is categorized as
  African-American and $Z=0$ otherwise.
  It contains 9.4K observations, and we retain 20\% of
  those for testing, and the rest for training/validation. We note that our
  preprocessing differs from the one used by \citet{Donini2018}: in
  particular, allowing us to
  retain more data. For completeness, we present in
  Appendix~\ref{subsec:additional-results-compas} the results obtained with
  the same preprocessing as \citet{Donini2018}.
    \item The \textit{Adult Income Dataset} (Adult), featured in
  \citep{Zafar2019,Donini2018}, is based on US census data and consists
  in predicting whether income exceeds \$50K a year. The sensitive
  variable is the gender of the individual, \textit{i.e.} male ($Z=1$) or
 female ($Z=0$).
  It contains 32.5K observations for training and validation, as well as
  16.3K observations for testing. For simplicity, we removed the weights
  associated to each instance of the dataset.
    \item The \textit{German Credit Dataset} (German), featured in
	\citep{Zafar2019,Zehlike2017FAIRAF,Singh2019,Donini2018}, 
	consists in classifying people described by a set of attributes as good
	or bad credit risks. The sensitive variable is the gender of the
	individual, \textit{i.e.} male ($Z=1$) or female ($Z=0$). It contains 1,000
 instances
	and we retain 30\% of those for testing, and the rest for
	training/validation.
    \item The \textit{Bank Marketing Dataset} (Bank), featured in
	\citep{Zafar2019}, consists in predicting whether a client will
	subscribe to a term deposit.  The sensitive variable is the age of the
	individual: $Z=1$ when the age is between 25 and 60 (which we refer to as
  ``working age population'') and $Z=0$ otherwise. It
	contains 45K observations, of which we retain 20\% for testing,
	and the rest for training/validation.
\end{itemize}
For all of the datasets, we used one-hot encoding for any categorical
variables. The number of training instances $n+m$, test instances $n_
\text{test}$
and features $d$ for each dataset is summarized in \cref{tab:data-dimension}.

\begin{table}[h]
    \centering
    \caption{Number of observations and feat $d$ per dataset.
    }\label{tab:data-dimension}
    \begin{tabular}{lcccccc}
	\toprule
	Dataset   & German    & Adult  & Compas    &  Bank  \\
	\midrule
	$n+m$ & 700 & 32.5K & 7.5K & 36K \\
	$n_\text{test}$  & 300 & 16.3K & 1.9K & 9K  \\
	$d$ & 61 & 107 & 16 & 59 \\
	\bottomrule
    \end{tabular}
\end{table}

\textbf{Parameters.}
For \cref{alg:sec3}, we select different $\auc$-based fairness constraints
for each dataset depending on the semantic of the task.
In the case of \textit{Compas} (recidivism prediction),
being labeled positive is a disadvantage so the approach with AUC-based
fairness uses the constraint in \cref{rk-cons:BPSN-AUC} to balance FPRs (by
forcing the probabilities that a negative from a given group is mistakenly
ranked higher than a positive to be the same across groups).
Conversely for \textit{German} (credit scoring), a positive label is an advantage, 
so we choose \cref{rk-cons:pairwise-accuracy} to balance FNRs.
For \textit{Bank} and
\textit{Adult}, the problem has no clear connotation so we select
\cref{rk-cons:inter-group-pairwise-fairness} to force the same ranking accuracy when comparing the positives of a group with the negatives of another.

Inspired by the consideration that many operational settings focus on
learning a good score for small FPR rates,
the ROC-based approach is configured to simultaneously
align the distribution of FPR and TPR for low FPRs
between both groups by penalizing solutions with
high $|\Delta_{H, 1/8}(s)|$, $|\Delta_{H, 1/4}(s)|$,
$|\Delta_{G, 1/8}(s)|$ and $|\Delta_{G, 1/4}(s)|$.

\begin{figure}[t]
     \centering

    \includegraphics[width=\columnwidth]{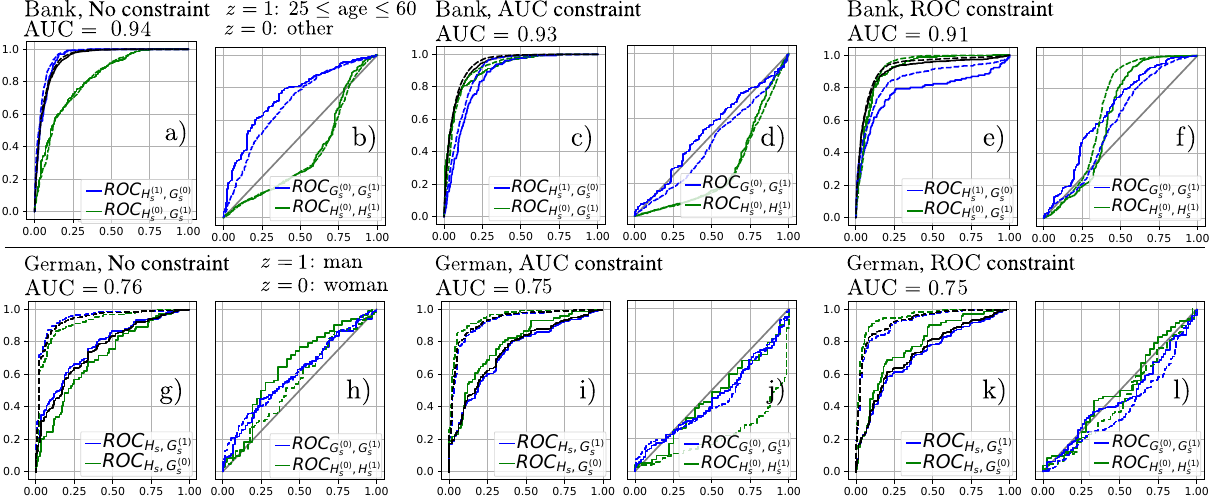}
    \caption{$\roc$ curves for Bank and German for a score learned without
    and with fairness constraints. On all plots, dashed and solid lines
    represent respectively training and test sets. Black curves represent
    $\roc_{H_s,G_s}$, and above the curves we report the
    corresponding ranking performance $\auc_{H_s,G_s}$.}\label{fig:roc_real_appendix}
	\bigskip
\end{figure}

\begin{figure*}
\centering
\includegraphics[width=\textwidth]{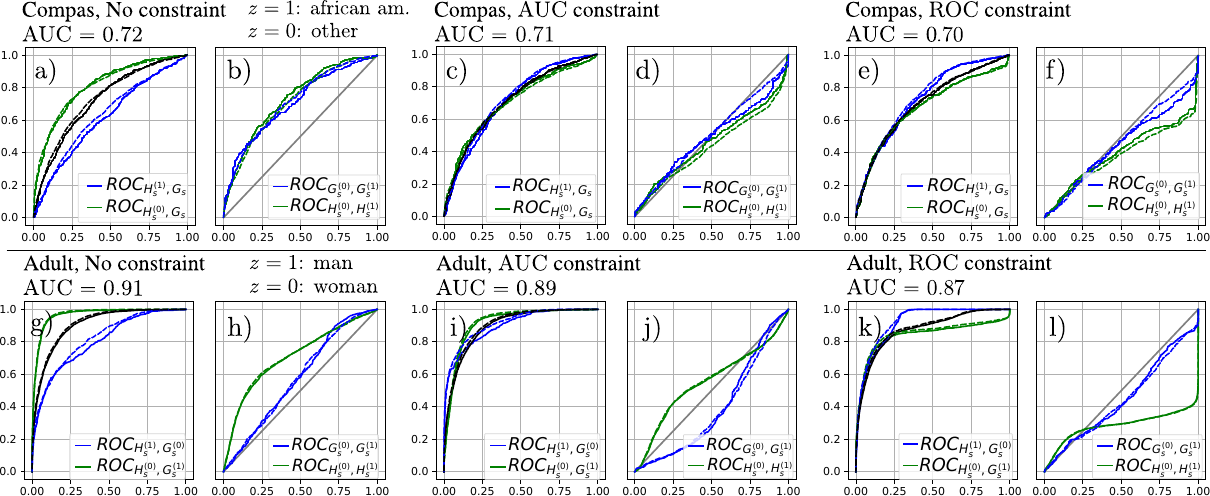}
    \caption{$\roc$ curves for Adult and Compas for a score learned without
    and with fairness constraints.
    On all plots, dashed and solid lines
    represent respectively training and test sets. 
    Black curves represent $\roc_{H_s,G_s}$, and above the curves
    we report the corresponding ranking performance $\auc_{H_s,G_s}$.
}\label{fig:roc_real_main_text2}
\end{figure*}

Precisely, for every run of \cref{alg:sec4}, we set:
\begin{align*}
    & m_G = m_H=2, \quad 
    \alpha_G^{(1)} = \alpha_H^{(1)} = \frac{1}{8}, \quad 
    \alpha_G^{(1)} = \alpha_H^{(2)} = \frac{1}{4}, \\
 & \lambda_G^{(1)} = \lambda_G^{(2)} = \lambda \quad  \text{and} \quad
 \lambda_H^{(1)} = \lambda_H^{(2)} = \lambda.
\end{align*}
For all algorithms, we chose the parameter
$\lambda$ from the candidate set $\in \{ 0, 0.25, 0.5, 1, 5, 10 \}$,
where $\lambda = 0$ corresponds to the case without constraint.
Denoting by $\widetilde{s}$ the output of \cref{alg:sec3} or \cref{alg:sec4},
we selected the parameter $\lambda_\text{reg}$ of the L2 regularization that
maximizes the criterion
$L_\lambda(\widetilde{s})$ (resp. $L_\Lambda(\widetilde{s})$) on the validation dataset
over the following candidate regularization strength set:
\begin{align*}
    \lambda_{\text{reg}} \in \{ 
	1 \times 10^{-3}, 
	5 \times 10^{-3},
	1 \times 10^{-2}, 
	5 \times 10^{-2}, 
	1 \times 10^{-1},
	5 \times 10^{-1},
	1 
    \}.
\end{align*}
The selected parameters are summarized in \cref{tab:table-all-parameters}.
Results are summarized in \cref{tab:table-all-results}, where $\auc$ denotes the ranking
accuracy $\auc_{H_s, G_s}$, and $\Delta \auc$ denotes the absolute difference
of the terms in the $\auc$-based fairness constraint of interest.
We also report on the values of $\abs{\Delta_{F, 1/8}}$ and $\abs{\Delta_{F,
1/4}}$ 
for $F \in \{ H, G \}$ and refer the reader to the $\roc$ curves in 
\cref{fig:roc_real_appendix} and \cref{fig:roc_real_main_text2} for
a visual summary of the other values of $\Delta_{F, \alpha}$ 
with $F \in \{H, G \}$ and $\alpha \in [0,1]$.
We highlight in bold the best ranking accuracy, and the
fairest algorithm for the relevant constraint.
All of the numerical evaluations reported below are evaluations on the
held-out test set.

\begin{table*}[h]
    \centering
    \caption{Parameters selected using the validation set
	for the runs on real data.
      }\label{tab:table-all-parameters}
      \begin{tabular}{lccccc}
	\toprule
	\multicolumn{2}{c}{Parameters} & \multicolumn{3}{c}{Constraint} \\
	\cmidrule(lr){1-2}\cmidrule(lr){3-5}
	 Dataset & Variable & None & $\auc$ & $\roc$ \\
	 \midrule 
	 \multirow{2}{*}{German} 
	 & $\lambda$ & 0 & 0.25 & 0.25\\
	 & $\lambda_\text{reg}$ & 0.5 & 0.5 & 0.5 \\
	\midrule
	 \multirow{2}{*}{Adult} 
	 & $\lambda$ & 0 & 0.25 & 0.25\\
	 & $\lambda_\text{reg}$ & 0.05 & 0.05 & 0.05 \\
	\midrule
	\multirow{2}{*}{Compas}
	 & $\lambda$ & 0 & 0.5 & 0.25\\
	 & $\lambda_\text{reg}$ & 0.05 & 0.05 & 0.05 \\
	\midrule
	\multirow{2}{*}{Bank}
	 & $\lambda$ & 0 & 0.25 & 0.25\\
	 & $\lambda_\text{reg}$ & 0.05 & 0.05 & 0.05 \\
	\bottomrule
      \end{tabular}
\end{table*}


\begin{table*}[t]
    \centering
    \caption{Results on test set. The strength of fairness constraints and
	regularization is chosen based on a validation set to obtain interesting
	trade-offs, as detailed in \cref{sec:appendix:real_data_experiments}. 
    }\label{tab:table-all-results}
    { 
    \begin{tabular}{llcccc}
	\toprule
	\multicolumn{2}{c}{Measure} 
	& \multicolumn{4}{c}{Dataset} \\
	\cmidrule(lr){1-2}\cmidrule(lr){3-6}
	Constraint & Value & German & Adult & Compas & Bank \\
	 \midrule 
	 \multirow{6}{*}{None}
	 &  $\auc$ & \textbf{0.76} & \textbf{0.91} & \textbf{0.72} & \textbf{0.94} \\
	 & $\Delta \auc$ & 0.07 & 0.16 & 0.20 & 0.13 \\
	 & $|\Delta_{H, 1/8}|$ & \textbf{0.01} & 0.31 & 0.26 & 0.09\\
	 & $|\Delta_{H, 1/4}|$ & 0.20  & 0.36 & 0.32 & 0.18\\
	 & $|\Delta_{G, 1/8}|$ & 0.13 & 0.02 & 0.29 & \textbf{0.00} \\
	 & $|\Delta_{G, 1/4}|$ & 0.20 & 0.06 & 0.29 & \textbf{0.04} \\
	 \midrule 
	 \multirow{6}{*}{$\auc$-based}
	  & $\auc$ & 0.75 & 0.89 & 0.71 & 0.93 \\
	 & $\Delta \auc$ & \textbf{0.05} & \textbf{0.02} & \textbf{0.00} & \textbf{0.05}\\
	 & $|\Delta_{H, 1/8}|$ & 0.05 & 0.09 & 0.06 & \textbf{0.03}\\
	 & $|\Delta_{H, 1/4}|$ & 0.08 & 0.17 & 0.03 & 0.11 \\
	 & $|\Delta_{G, 1/8}|$ & \textbf{0.01} & 0.06 & 0.02 & 0.27\\
	 & $|\Delta_{G, 1/4}|$ & 0.02 & 0.14 & 0.06 & 0.37\\
	 \midrule 
	 \multirow{6}{*}{$\roc$-based}
	 & $\auc$ & 0.75 & 0.87 & 0.70 & 0.91 \\
	 & $\Delta \auc$ & 0.07 & 0.07 & 0.05 & 0.14 \\
	 & $|\Delta_{H, 1/8}|$ & 0.03 & \textbf{0.06} & \textbf{0.01} & \textbf{0.03} \\
	 & $|\Delta_{H, 1/4}|$ & \textbf{0.07} & \textbf{0.01} & \textbf{0.02} & \textbf{0.05} \\
	 & $|\Delta_{G, 1/8}|$ & 0.04 & \textbf{0.00} & \textbf{0.00} & 0.06 \\
	 & $|\Delta_{G, 1/4}|$ & \textbf{0.01} & \textbf{0.02} & \textbf{0.00} & 0.21 \\
	\bottomrule
    \end{tabular}
    }
\end{table*}

\textbf{Results for the datasets \emph{Compas} and \emph{Adult}.} These
results are presented in Section~\ref{sec:experiments} of the main text.
For completeness, \cref{fig:roc_real_main_text2}
represents $\roc$ curves on the training set of \emph{Compas} and \emph{Adult},
on top of the test set $\roc$ curves already displayed in 
\cref{fig:roc_real_main_text} of the main text.

\textbf{Results for the dataset \textit{Bank}.}
Recall that for this dataset we consider the $\auc$ constraint
\cref{rk-cons:inter-group-pairwise-fairness} to force the same ranking
accuracy when comparing the positives of a group with the negatives of another.
\cref{fig:roc_real_appendix} shows that the score learned without constraint
implies a stochastic order between the distributions of the problem that writes 
$H_s^{(1)} \preceq  H_s^{(0)} \preceq G_s^{(0)} \preceq G_s^{(1)}$, where
$h \preceq g$ means that $g$ is stochastically larger than $h$
(Fig. \ref{fig:roc_real_appendix}-b).
This suggests that the task of distinguishing positives from negatives is much
harder for observations of the group $Z = 0$ than for those of the working age
population ($Z=1$),
which could be a consequence of the heterogeneity of the group $Z=0$.
On the other hand, the left plot representing $\roc_{H_s^{(1)},G_s^{(0)}}$
and $\roc_{H_s^{(0)},G_s^{ (1)}}$ (Fig. \ref{fig:roc_real_appendix}-a)
for the setting without constraint gives an appreciation
of the magnitude of those differences. Precisely, it implies that it is much
harder to distinguish working age positives ($Y=+1, Z=1$) from negatives of
group $Z=0$ than working age negatives from positives of group $Z=0$
(Fig. \ref{fig:roc_real_appendix}-a).
The correction induced by the $\auc$ constraint suggests that it was due to
the fact that scores for positives of the group ($Y=+1, Z=0$) were too small compared
to the positives of the working age population ($Y=+1, Z=1$). Indeed, learning
with the $\auc$ constraint roughly equalizes the scores of the positives
across both groups $Z=0$ and $Z=1$ (Fig. \ref{fig:roc_real_appendix}-d).
Additionally, in the left plot for learning with $\auc$
constraints, we can see that $\roc_{H_s^{(1)},G_s^{(0)}}$ and $\roc_
{H_s^{(0)},G_s^{
(1)}}$ intersect and have similar $\auc$'s as expected (Fig. \ref{fig:roc_real_appendix}-c),
which is more visible for the dashed lines (\textit{i.e.} on training data).
Finally, the $\roc$-based constraint induces as expected the equality of 
$G_s^{(0)}$ and $G_s^{(1)}$ as well as that of $H_s^{(0)}$ and $H_s^{(1)}$
in the high score regime, as seen on the right plot (Fig. \ref{fig:roc_real_appendix}-f).
It implies that $\roc_{H_s^{(1)},G_s^{(0)}}$
and $\roc_{H_s^{(0)},G_s^{(1)}}$ are much closer for simultaneously small TPR's and FPR's
(Fig. \ref{fig:roc_real_appendix}-e),
which entails that thresholding top scores will yield fair
classifiers in FPR and TPR again for a whole range of high thresholds.

\textbf{Results for the dataset \textit{German}.}
Recall that for this credit scoring dataset we consider
the $\auc$-based constraint in \cref{rk-cons:pairwise-accuracy} to force the probabilities
that a positive from a given group is mistakenly ranked higher than a negative to be
the same across groups.
Despite the blatant issues of generalization due to the very small
size of the dataset (see \cref{tab:data-dimension}), we see in 
\cref{fig:roc_real_appendix}
that the learned score without fairness constraints systematically makes more
errors for women with good ground truth credit risk, as can be seen from
comparing $\roc_{H_s,G_s^{(0)}}$
and $\roc_{H_s,G_s^{ (1)}}$
(Fig. \ref{fig:roc_real_appendix}-g)
Additionally, the credit score of men with good or bad credit risk is in both
cases stochastically larger than that of women of the same credit risk assessment
(see $\roc_{G_s^{(0)},G_s^{(1)}}$ and $\roc_{H_s^{(0)},H_s^{(1)}}$
in Fig. \ref{fig:roc_real_appendix}-h).
On the other hand, the score learned with an $\auc$ constraint makes
a similar amount of mistakes for both genders, with only slightly more mistakes
made on men than women (Fig. \ref{fig:roc_real_appendix}-i),
and the scores $s(X)$ conditioned on the events $(Y=y, Z=z)$
with $z = 0$ and $z=1$ are more aligned when considering both $y=-1$ and $y=+1$
(Fig. \ref{fig:roc_real_appendix}-j).
Finally, while the score learned with a $\roc$ constraint has a slightly
higher discrepancy between the $\auc$'s involved in \cref{rk-cons:pairwise-accuracy}
than the one learned with an $\auc$ constraint (Fig. \ref{fig:roc_real_appendix}-k),
one observes that both pairs of distributions $(G_s^{(0)}, G_s^{(1)})$ and
$(H_s^{(0)}, H_s^{(1)})$ are equal for high thresholds (Fig. \ref{fig:roc_real_appendix}-l).
Consistently with the results on other datasets, this suggests that our
score leads to classifiers that are fair in FPR and TPR for a whole range of
problems where one selects individuals with very good credit risks by
thresholding top scores.

\subsection{Results on Compas with a Different
Preprocessing}\label{subsec:additional-results-compas}

\begin{figure*}
\centering
\includegraphics[width=\textwidth]{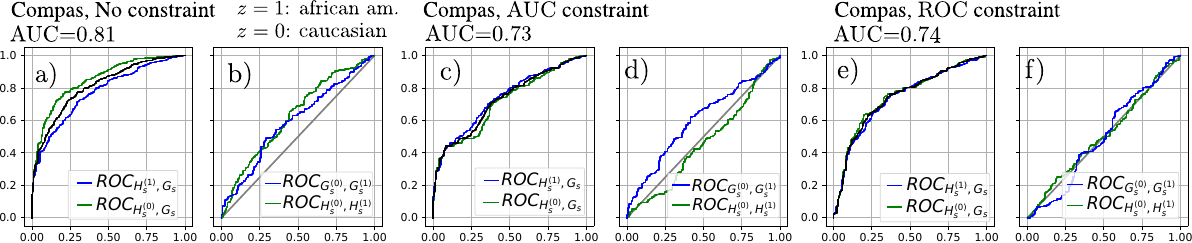}
    \caption{$\roc$ curves for Compas for the preprocessing
	described in \cite{Donini2018},
	and a score learned without
    and with fairness constraints.
    On all plots, dashed and solid lines
    represent respectively training and test sets. 
    Black curves represent $\roc_{H_s,G_s}$, and above the curves
    we report the corresponding ranking performance $\auc_{H_s,G_s}$.
}\label{fig:roc_compas_only_new}
\end{figure*}

For the dataset \textit{Compas}, the results in \cref{sec:experiments}
use a different preprocessing from \cite{Donini2018} which allows us to
retain more data.
For completeness, we perform additional experiments
using the same data and preprocessing as \cite{Donini2018}.
The main difference is that the sensitive variable to be $Z= 1$ if
an individual is
categorized as African-American and $Z=0$ if it is categorized as Caucasian 
(observations associated with other ethnicities are discarded).
The dataset then contains 5.3K observations ($n+m$), and we retained 20\%
of those for testing, and the rest for training/validation.
Results are reported in \cref{fig:roc_compas_only_new}.

\end{document}